\documentclass[11pt]{article}
\usepackage[preprint]{acl}
\usepackage{times}
\usepackage{latexsym}
\usepackage[T1]{fontenc}
\usepackage[utf8]{inputenc}
\usepackage{microtype}
\usepackage[noupquote]{inconsolata}
\usepackage{graphicx}
\usepackage{subcaption}
\usepackage{booktabs}
\usepackage{amsmath}
\usepackage{array}
\usepackage{xcolor}
\usepackage{listings}
\usepackage{placeins}
\usepackage{float}
\graphicspath{{figures/}{figures/main/}{figures/supplement/}}
\hypersetup{
  pdftitle={LLMs Can Predict Failure Risk, But Struggle to Predict Which Collaboration Protocol Pays Off: Cost-Aware Protocol Routing Across Reasoning Tasks},
  pdfauthor={Chih-Hsuan Yang, Jingyan Jiang, Cheng-Hau Yang, Vikram Vasudevan, Huihuo Zheng, Venkatram Vishwanath, Rajeev Thakur},
  pdfsubject={Cost-aware protocol routing for LLM reasoning},
  pdfkeywords={LLM agents, multi-agent systems, cost-aware routing, protocol routing, calibration}
}

\title{LLMs Can Predict Failure Risk, But Struggle to Predict Which
Collaboration Protocol Pays Off:\\
Cost-Aware Protocol Routing Across Reasoning Tasks}

\author{
\textbf{Chih\mbox{-}Hsuan Yang$^{1}$,
Jingyan Jiang$^{1}$,
Cheng\mbox{-}Hau Yang$^{1}$,
Vikram Vasudevan$^{2}$,}\\
\textbf{Huihuo Zheng$^{1}$,
Venkatram Vishwanath$^{1}$,
Rajeev Thakur$^{1}$}\\
$^{1}$Argonne National Laboratory, Lemont, IL, USA\\
$^{2}$Oregon State University, Corvallis, OR, USA\\
\texttt{bellayang@anl.gov}
}

\begin{document}
\maketitle

\begin{abstract}
Multi-agent large language model (LLM) systems can improve reasoning by
spending more computation, but deployment requires deciding when extra
collaboration is worth its cost. We isolate this decision by running every
problem under four protocols while holding the solver fixed within each
setting: direct solving (Baseline), iterative self-correction (Single),
planner--executor--reviewer collaboration (PER), and multi-agent deliberation
(Broadcast). The primary benchmark comprises 4{,}181 competition-level math problems;
paired robustness checks cover four benchmarks spanning competition math,
biology, and broader science with two solver families.
Across fixed policies, trained routers, and frozen LLM routers, conservative
policies under-escalate, whereas higher-solve frozen routers often
over-escalate. A post-answer, pre-collaboration \texttt{gpt-oss-120b} probe
ranks Baseline failures with 0.8847 AUROC (4{,}151 parseable cases; 95\% CI
[0.8732, 0.8955]). The same score remains informative for predicting whether any
collaboration helps (0.7683 AUPRC), but is much weaker for identifying PER- or
Broadcast-specific value (0.1674 and 0.1041 AUPRC). Separately, the pre-answer
self-confidence gate reaches 78.0\% solve at 45K tokens, compared
with 73.8\% at 71.3K for a frozen \texttt{gpt-oss-120b} router and 92.4\% for
a retrospective fixed-order oracle. Across 10 paired model--condition
settings, the oracle adds 23.2--58.3 points of retrospective coverage over
Baseline, but protocol profiles vary by task. In the six settings with
held-out router evaluations, oracle gaps remain 18.5--28.9 points. Confidence
can therefore support initial escalation, while protocol-specific cost-aware
routing remains unresolved.
\end{abstract}

\section{Introduction}\label{sec:intro}

Large language model (LLM) reasoning systems can spend drastically different
amounts of computation on the same problem. A direct solver may cost tens of
thousands of tokens, while self-correction or multi-agent deliberation can cost
an order of magnitude more. Stronger protocols are often more accurate, but
deploying the strongest protocol everywhere silently multiplies cost.
Deployment therefore requires deciding both \emph{whether} to escalate and
\emph{which} collaboration protocol is worth its marginal cost.

We build a matched benchmark of 4{,}181 competition-level math problems, each
evaluated under four protocols with sharply different token costs and solve
rates. Observing every protocol on every problem permits offline evaluation
against realized solve and cost outcomes and a retrospective fixed-order
oracle. This design reveals a directional failure hidden by routing accuracy:
cost-conservative policies under-escalate, whereas higher-solve frozen LLM
routers often over-escalate.

These observations motivate a sharper distinction. \emph{Failure-risk
prediction} asks whether Baseline will be wrong; \emph{collaboration-value
prediction} asks whether, and which, stronger protocol justifies its added
cost. A full-benchmark post-answer probe directly supports the first claim, but
its precision degrades for increasingly protocol-specific value targets.
Matched outcome checks extend across four benchmarks spanning competition
math, biology, and broader science, using both \texttt{gpt-oss-120b} and
\texttt{Gemma-4-31B-it}
\citep{gemma4report,laurent2024labbench,arora-etal-2023-jeebench,wang2024scibench}.
The more complete confidence and held-out-router evaluations remain a targeted
six-setting subset, so this breadth is evidence of task dependence rather than
universality.

Several related literatures vary different inference components. Model routing
selects among models or cascades with different capabilities and prices
\citep{chen2023frugalgpt,ong2024routellm,somerstep2025carrot,song-etal-2025-irt}.
Adaptive computation and sample routing vary reasoning depth or the number of
sampled paths \citep{wang2023selfconsistency,balachandran2025inference,wu-etal-2025-thought}.
Tool and multi-agent routing vary tools, roles, or collaboration structures
\citep{yue-etal-2025-masrouter,gan-etal-2025-master,zhu-etal-2025-multiagentbench}.
Calibration work instead asks whether confidence tracks correctness
\citep{joshi-etal-2025-calibration,li-etal-2025-large-language-models,li-etal-2025-task}.
Our setting holds the solver family fixed within each comparison and changes
only the collaboration protocol. This isolates protocol value from a change in
base model capability; Supplementary Table~S1 summarizes these distinctions.

\paragraph{Contributions.}
First, we provide matched four-protocol outcomes and solve--cost evaluation that
keep the solver fixed within each setting, with paired coverage checks across
four benchmarks. Second, we identify asymmetric under- versus over-escalation across
heuristic, learned, and frozen-LLM routers. Third, we empirically separate
failure risk from protocol-specific collaboration value: same-model confidence
supports an initial stay-or-escalate decision, while substantial
fixed-order-oracle gaps and weak protocol-specific precision leave full
cost-aware routing open. The source package includes machine-readable aggregate
tables, and the companion dataset archive contains anonymized protocol traces
and outcome labels for the matched comparisons.

\section{Task, Benchmark, and Metrics}
\label{sec:task}

\paragraph{Protocol routing.}
A \emph{router} selects one action before observing protocol outcomes. Fixed
policies are degenerate routers; learned and frozen-LLM routers map problem
text and/or allowed metadata to \textsc{Baseline}, \textsc{Single},
\textsc{PER}, \textsc{Broadcast}, or \texttt{None}. The \texttt{None} action
abstains and can be both an oracle label and a router prediction.

Our main benchmark uses the clean exact-answer Omni-MATH~2 subset, a manually
revised release derived from Omni-MATH, with 4{,}181 competition-level math
problems \citep{omnimath,omnimath2}. Router-visible metadata includes source,
domain path, numeric difficulty, and a ten-level difficulty tier; real examples
include source values \texttt{cayley}, \texttt{fermat}, and \texttt{pascal}; one
domain path is
\texttt{Mathematics -> Algebra -> Prealgebra -> Simple Equations}. For the
LAB-Bench robustness study, visible fields are dataset name, domain,
slice/subset, original identifier, prompt condition, and subtask, e.g.,
\texttt{Biology -> LAB-Bench -> CloningScenarios}. Routers and confidence
probes never receive gold answers, correctness labels, oracle labels, or
protocol outcomes.

\paragraph{Protocols.}
In the main benchmark, every problem is run once under four protocols with the
same \texttt{gpt-oss-120b} solver stack \citep{openai_gptoss_model}.
\textsc{Baseline} is one direct attempt
without self-correction. \textsc{Single} adds iterative self-correction.
\textsc{PER} uses planner, executor, and reviewer roles. \textsc{Broadcast}
uses multi-agent deliberation with shared candidates and peer approval. On the
main held-out split, these protocols average 18.2K, 47.6K, 401.9K, and 622.1K
tokens and solve 56.3\%, 78.5\%, 84.9\%, and 88.9\%, respectively. Runs use
temperature 0.0; protocol definitions and compute accounting are detailed in
the supplement.

\paragraph{Fixed-order oracle.}
For each problem, the retrospective oracle is the first successful action in
the aggregate cost order \textsc{Baseline} $<$ \textsc{Single} $<$
\textsc{PER} $<$ \textsc{Broadcast}; it is \texttt{None} if all four fail.
This is a matched, single-realization diagnostic upper bound, not a deployable
policy, a per-instance minimum-token oracle, or expected success under repeated
sampling. On the full benchmark, its labels are 56.8\% Baseline, 23.0\%
Single, 8.8\% PER, 4.2\% Broadcast, and 7.3\% None.

\begin{figure*}[!t]
\centering
\includegraphics[width=\linewidth]{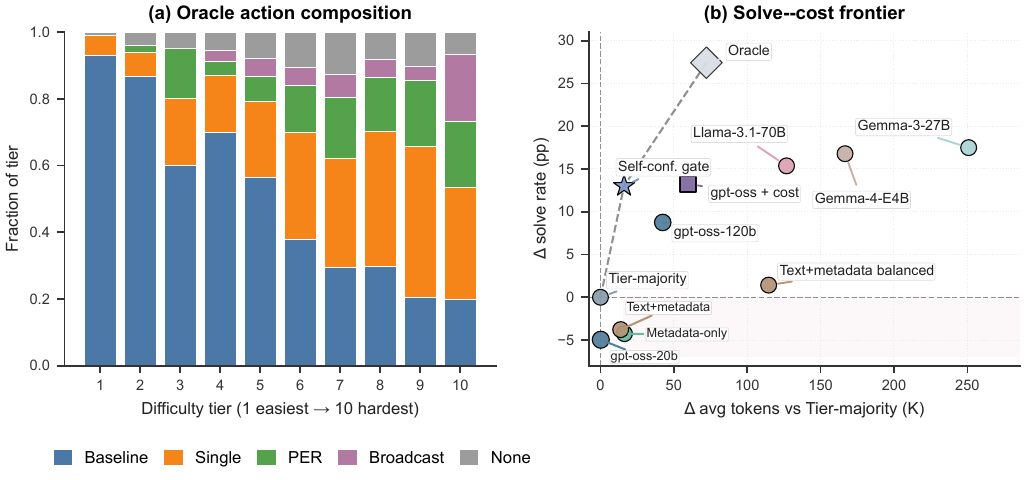}
\caption{(a) Fixed-order-oracle labels by difficulty tier; fractions sum to 1
within each tier. (b) Solve--cost frontier relative to Tier-majority; upper-left
is better. The self-confidence gate improves the low-cost frontier, while
larger frozen routers buy additional solves through substantially more token
use. Gemma-4-E4B here is a frozen router over the main \texttt{gpt-oss} traces,
not the Gemma-4-31B-it solver used in Section~\ref{sec:robustness}.}
\label{fig:motivation-tradeoff}
\end{figure*}

\begin{table*}[!t]
\centering
\small
\setlength{\tabcolsep}{5.2pt}
\begin{tabular}{llrrr}
\toprule
Class & Policy & Solve (\%) [95\% CI] & Avg. tokens (K) [95\% CI] & Excess (K) [95\% CI] \\
\midrule
Fixed & Baseline & 56.3 [51.5, 61.0] & 18.2 [17.1, 19.2] & 1.7 [1.1, 2.4] \\
Heuristic & Tier-majority & 65.0 [60.5, 69.5] & 28.9 [26.0, 31.9] & 5.7 [3.9, 7.7] \\
Frozen LLM & \texttt{gpt-oss-120b} & 73.8 [69.5, 77.5] & 71.3 [56.5, 86.7] & 37.1 [24.2, 50.6] \\
Frozen LLM & \texttt{gpt-oss} + cost prompt & 78.3 [74.2, 82.0] & 88.6 [71.0, 107.6] & 51.6 [35.2, 68.9] \\
Confidence & Self-confidence gate & 78.0 [74.0, 81.8] & 45.0 [41.6, 48.7] & 14.8 [12.2, 17.4] \\
Retrospective & Fixed-order oracle & 92.4 [89.8, 95.0] & 101.1 [80.1, 124.5] & 0.0 [0.0, 0.0] \\
\bottomrule
\end{tabular}
\caption{Representative routing policies on the primary held-out test split
($n=423$). Intervals use 2{,}000 problem-level bootstrap resamples. Excess is
positive token overpayment above the realized fixed-order oracle. The protocol
executions and router/probe calls use deterministic decoding, so the intervals
do not measure fresh-run variability.}
\label{tab:main-routing}
\end{table*}

\paragraph{Routers and evaluation.}
The main learned-router comparison uses a stratified 80/10/10 split (seed 42;
3{,}342/416/423 problems). \emph{Tier-majority} predicts each difficulty tier's
train-split majority fixed-order-oracle label, falling back to the train-split
global majority. It is a transparent metadata-only sanity check, not a deployed
heuristic.
Learned routers are five-class logistic regressions over allowed metadata, with
optional TF--IDF word uni- and bigrams from the problem text; model and
hyperparameter selection use the development split only. Frozen routers receive
the same text and metadata in a label-only prompt. The cost-aware ablation also
provides numeric average protocol costs and short routing examples.

We report solve rate, average tokens, and \emph{excess tokens}, the mean positive
per-problem token overpayment relative to the realized oracle. Under- and
over-escalation mean choosing cheaper and costlier actions than the oracle.
\texttt{None} has zero protocol cost and ranks below Baseline; router or probe
overhead still counts. Hence, choosing a protocol when the oracle is
\texttt{None} is over-escalation, and all its tokens are excess.

\section{Routing Errors and Cost Tradeoffs}
\label{sec:results}

We compare fixed and heuristic policies, lightweight trained routers, frozen
LLM routers, a cost-aware prompt, and the confidence policy introduced in
Section~\ref{sec:mechanism}. Figure~\ref{fig:motivation-tradeoff} shows the
oracle composition by tier and the resulting solve--cost frontier. Table
\ref{tab:main-routing} reports representative policies with uncertainty; the
supplement gives the full grouped comparison.

\paragraph{Cheap references are competitive.}
Baseline solves 56.3\% at 18.2K average tokens. Tier-majority raises solve rate
to 65.0\% at 28.9K tokens. The selected metadata-only and text+metadata
logistic routers solve 60.8\% and 61.2\%, respectively, below this simple
reference. Balanced and embedding variants recover more solves only by routing
more often to expensive actions; full results and feature ablations appear in
the supplement.

\paragraph{Frozen routers trade cost for fewer misses.}
The frozen \texttt{gpt-oss-120b} router reaches 73.8\% solve at 71.3K tokens.
Showing numeric protocol costs and routing examples raises it to 78.3\%, but
also to 88.6K tokens and 51.6K excess. Larger cross-family frozen routers reach
80--83\% solve while spending 156K--280K tokens on average. No evaluated router approaches
the fixed-order-oracle operating point.

\paragraph{Errors are directional.}
Tier-majority under-escalates on 27.4\% and over-escalates on 12.5\% of test
problems. The \texttt{gpt-oss-120b} router cuts under-escalation to 18.0\% but
raises over-escalation to 33.3\%. Higher-solve Llama and Gemma frozen routers
reduce under-escalation to 6--11\% while over-escalating on 63--71\%. Thus,
router gains are not interchangeable: conservative policies miss recoverable
problems; aggressive routers buy solves through costly escalation.

\FloatBarrier

\section{Failure Risk Is Not Protocol Value}
\label{sec:mechanism}

The directional errors raise two different questions. Can the solver recognize
that its Baseline answer is likely wrong? If so, can it identify which stronger
protocol is worth the added cost?

\paragraph{Post-answer failure-risk probe.}
After \texttt{gpt-oss-120b} produces its Baseline answer, we ask for
$P(\text{Baseline correct})$. The probe sees only the problem, allowed
metadata, and its own Baseline \emph{final answer}; it sees no reasoning trace,
gold answer, correctness label, oracle label, or collaboration outcome. Of
4{,}181 matched problems, 4{,}151 produce parseable scores (99.28\%). The 30
unparseable outputs are excluded rather than imputed. Using
$1-P(\text{correct})$ as failure risk yields 0.8847 AUROC (95\% CI [0.8732,
0.8955]), 0.8950 AUPRC, and 0.0852 expected calibration error.

This differs from the \emph{pre-answer confidence probe}, which asks for
single-pass solve probability before the model sees a Baseline answer. That
probe has 0.859 AUROC on 329 usable estimates from the primary 423-example test split.
The two results correspond to distinct operating points and are not treated as
replicates.

\begin{table}[b]
\centering
\scriptsize
\setlength{\tabcolsep}{3.0pt}
\begin{tabular}{lrrr}
\toprule
Target & Prev. & AUROC & AUPRC \\
\midrule
Baseline fails & 43.4 & 0.8847 & 0.8950 \\
Any collaboration helps & 36.0 & 0.8544 & 0.7683 \\
PER first success & 8.8 & 0.7259 & 0.1674 \\
Broadcast-only success & 4.2 & 0.7639 & 0.1041 \\
\bottomrule
\end{tabular}
\caption{The same post-answer failure score on increasingly specific targets
($n=4{,}151$ parseable cases). Prevalence is a percentage. ``PER first'' means
Baseline and Single fail but PER succeeds; ``Broadcast-only success'' means
Baseline, Single, and PER fail while Broadcast succeeds in the matched realized
runs.}
\label{tab:value-targets}
\end{table}

\paragraph{Protocol-specific precision is much weaker.}
Table~\ref{tab:value-targets} applies the same no-leakage failure score to
increasingly specific collaboration-value targets. It remains useful for the
coarse question of whether any protocol improves on Baseline. For PER and
Broadcast, however, AUPRC falls to 0.1674 and 0.1041. These exceed the rare
target prevalences, so confidence is not devoid of value signal, but it is not
a reliable full protocol selector.

\paragraph{A binary gate is useful but narrower.}
The self-confidence gate uses the earlier pre-answer confidence score: keep
Baseline at confidence $\geq70$, otherwise choose Single; missing confidence scores also choose
Single. The threshold is selected on dev by a knee-point rule
\citep{satopaa2011finding,deb2011understanding}. On the full test split it
reaches 78.0\% solve at 45.0K tokens, versus 73.8\% at 71.3K for the frozen
\texttt{gpt-oss} router (Table~\ref{tab:main-routing}). The overlapping solve
intervals support a cost-efficiency comparison, not a claim of higher solve
rate. We reserve \emph{self-confidence gate} for this binary policy;
\emph{two-threshold cascade} denotes the separate supplementary ablation that
uses Baseline, Single, and Tier-majority with high/low thresholds of 70/10.

The gate does not choose PER, Broadcast, or None. Its 14.4-point gap to the
fixed-order oracle and the target-specific results identify the remaining
problem: estimating the marginal value of each expensive protocol after the
initial escalation decision.

\begin{figure*}[!t]
\centering
\includegraphics[width=\linewidth]{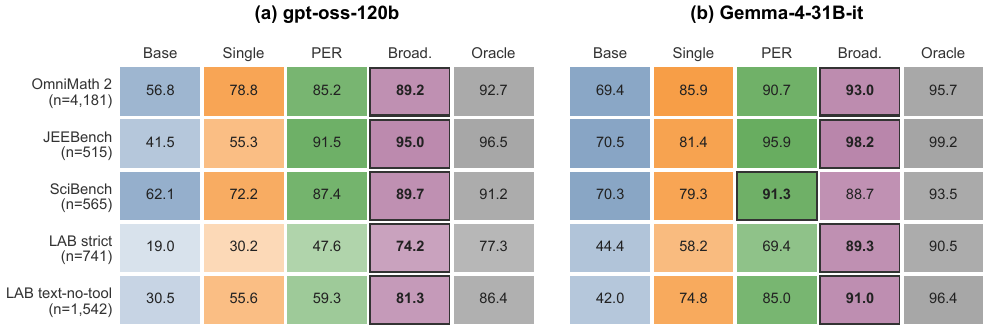}
\caption{Matched solve/coverage percentages across four benchmarks (five prompt
conditions), four protocols, and two solver families. Cell hue identifies the
protocol, and deeper tint indicates higher solve/coverage. The outlined cell
is the strongest deployable fixed protocol in each row. Oracle is retrospective
fixed-order coverage computed from the same four realized outcomes. Unlike
Table~\ref{tab:main-routing}, which reports the primary held-out test split,
this figure uses all matched problems available in each setting. Each value is
one deterministic realized execution per problem--protocol pair.}
\label{fig:matched-robustness}
\end{figure*}

\section{Matched Breadth and Router Robustness}
\label{sec:robustness}

To test whether the matched protocol pattern is confined to OmniMath, we extend
the analysis to four benchmarks across three broad task families: OmniMath~2 (competition math),
JEEBench (engineering entrance-exam STEM), SciBench (college-level science),
and LAB-Bench (biology multiple-choice).
LAB-Bench contributes two no-tool prompt conditions:
``strict'' includes relevant evidence in the prompt, while ``text-no-tool'' is
a broader text-only slice. Both \texttt{gpt-oss-120b} and
\texttt{Gemma-4-31B-it} cover all five conditions, yielding 10 paired settings.

Stronger collaboration has recoverable value in all 10 paired settings
(Figure~\ref{fig:matched-robustness}): fixed-order-oracle coverage exceeds
Baseline by 23.2--58.3 points. The protocol profile varies, however. Broadcast
is the strongest fixed protocol in nine settings, while PER exceeds Broadcast
by 2.7 points for Gemma on SciBench. This breadth supports task-dependent
protocol value, not universal generalization.

The deeper confidence and held-out-router analyses use the narrower
six-setting subset comprising both solvers on OmniMath and the two LAB-Bench
conditions. We train the same pre-specified text+metadata logistic router using
a stratified 70/15/15 split (seed 20260712), dev selection only, and refitting
on train+dev. It improves over Baseline by 7.2--37.5 points in all six settings,
yet remains 18.5--28.9 points below the fixed-order oracle. Relative to
Tier-majority, gains are mixed. Thus, observable text and metadata contain
routing signal but do not yield uniformly reliable protocol-value routing.

The expensive protocols are also not interchangeable. Conditional on Baseline
and Single both failing, Broadcast exceeds PER by 9.6--10.7 points on OmniMath
and 30.8--44.6 points on LAB-Bench. Nevertheless, PER-only successes remain in
every setting (2.4--11.1\% of these conditional subsets), so Broadcast does not
pointwise dominate PER.

\section{Discussion and Conclusion}
\label{sec:discussion}

The matched design separates two decisions that aggregate routing accuracy can
hide. Failure-risk estimation is useful for deciding whether to remain with a
cheap direct solver. Collaboration-value estimation must additionally predict
which expensive workflow provides enough marginal benefit. Conservative and
aggressive routers fail in opposite directions, the binary gate covers only
the first decision, and the broader held-out routers retain large oracle gaps.

Our conclusion is therefore diagnostic rather than a claim to have solved full
routing. Same-model confidence can support an initial stay-or-escalate gate,
but selecting among Single, PER, Broadcast, and None remains a cost-aware
control problem. The benchmark, directional error analysis, and explicit
failure-risk/protocol-value distinction provide measurable targets for that
next step.

\section{Limitations}
Our main router comparison and pre-answer confidence gate use one solver family
and one math benchmark. The broader paired matched-outcome study covers four
benchmarks across three broad task families and two solvers. Post-answer
confidence and held-out router analyses cover the narrower six settings formed
by both solvers on OmniMath and the two LAB-Bench conditions. These are
targeted robustness checks, not a claim of broad universality. Each
problem--protocol pair has one deterministic realized outcome. The
2{,}000-resample intervals quantify variation over benchmark problems, not
fresh-run stochastic variability. The fixed-order oracle is retrospective and
uses an aggregate cost order; it is neither deployable nor an estimate of
per-instance expected utility.

We measure cost primarily in logged tokens. Latency, monetary price, energy,
parallelism, and answer quality beyond exact correctness can change deployment
preferences. The post-answer probe excludes 30 unparseable outputs from its
probability metrics, and confidence quality varies substantially by model and
domain. Finally, the observed PER--Broadcast differences do not identify a
causal mechanism; understanding why protocol value changes by task remains
open.

\section*{Acknowledgments}
This research used resources of the Argonne Leadership Computing Facility, a
U.S. Department of Energy (DOE) Office of Science user facility at
\mbox{Argonne~National} Laboratory (ANL) operated under Contract
No.~DE-AC02-06CH11357.

\bibliography{custom}

\clearpage
\appendix
\setcounter{figure}{0}
\setcounter{table}{0}
\renewcommand{\thefigure}{S\arabic{figure}}
\renewcommand{\thetable}{S\arabic{table}}
\section{Cross-Setting Robustness and Diagnostics}
\label{app:additional-analyses}

This section establishes the scope of the main claims through targeted
robustness and diagnostic analyses. Unless stated otherwise, confidence intervals use
2{,}000 percentile bootstrap resamples over problem identifiers.

\subsection{Positioning Within Routing and Adaptive Inference}

Table~\ref{tab:routing-scope} makes explicit which inference component each
nearby line of work changes. Model routing and cascades select among base
models \citep{chen2023frugalgpt,ong2024routellm,somerstep2025carrot,song-etal-2025-irt};
adaptive computation and self-consistency allocate steps or samples
\citep{wang2023selfconsistency,balachandran2025inference,wu-etal-2025-thought};
tool and agent routing changes tools, roles, or collaboration structures
\citep{yue-etal-2025-masrouter,gan-etal-2025-master}. Our matched analysis
changes the collaboration protocol while holding the solver fixed within a
setting and observing every action's realized outcome.

\begin{table*}[t]
\centering
\scriptsize
\setlength{\tabcolsep}{3.2pt}
\begin{tabular}{lllll}
\toprule
Family & Routed unit & Fixed solver? & Matched outcomes? & Main target \\
\midrule
Model routing/cascades & Base model & No & Usually no & Quality--price tradeoff \\
Adaptive computation & Steps/depth & Usually & Usually no & Stop or continue \\
Self-consistency routing & Sampled reasoning paths & Yes & Usually no & Number of samples \\
Tool/agent routing & Tools, roles, topology & Varies & Usually no & Workflow selection \\
This work & Collaboration protocol & Yes & Yes & Realized solve--token value \\
\bottomrule
\end{tabular}
\caption{Scope comparison. ``Matched'' means that every candidate action is
observed for every evaluated problem, enabling paired offline analysis.}
\label{tab:routing-scope}
\end{table*}

\subsection{Matched Protocol Coverage Across Four Benchmarks}

We evaluate matched four-protocol outcomes on four benchmarks across three
broad task families: OmniMath~2 (competition math), JEEBench (engineering
entrance-exam STEM), SciBench (college-level science), and LAB-Bench (biology
multiple-choice)
\citep{omnimath2,arora-etal-2023-jeebench,wang2024scibench,laurent2024labbench}.
LAB strict is an LLM-only, no-tool/no-retrieval condition with relevant evidence
included in the prompt; LAB text-no-tool is a broader text-only no-tool
condition. Both \texttt{openai/gpt-oss-120b} \citep{openai_gptoss_model} and
\texttt{google/gemma-4-31B-it} \citep{gemma4report} cover all five evaluation
conditions. Tables \ref{tab:robustness-full-ci} and
\ref{tab:robustness-oracle-labels} therefore report 10 matched
model--condition settings.

\begin{table*}[t]
\centering
\scriptsize
\setlength{\tabcolsep}{2.7pt}
\begin{tabular}{lrccccc}
\toprule
Setting & $n$ & Baseline & Single & PER & Broadcast & Oracle \\
\midrule
\multicolumn{7}{l}{\textit{gpt-oss-120b}} \\
OmniMath~2 & 4{,}181 & 56.83 [55.4,58.4] & 78.78 [77.6,80.0] & 85.17 [84.1,86.2] & 89.21 [88.3,90.2] & 92.68 [91.9,93.5] \\
JEEBench & 515 & 41.55 [37.5,45.8] & 55.34 [51.1,59.6] & 91.46 [88.9,94.0] & 94.95 [93.0,96.7] & 96.50 [94.8,98.1] \\
SciBench & 565 & 62.12 [58.1,66.0] & 72.21 [68.5,75.8] & 87.43 [84.6,90.1] & 89.73 [87.1,92.2] & 91.15 [88.7,93.5] \\
LAB strict & 741 & 19.03 [16.1,21.9] & 30.23 [27.0,33.7] & 47.64 [44.0,51.3] & 74.22 [71.1,77.5] & 77.33 [74.2,80.3] \\
LAB text-no-tool & 1{,}542 & 30.54 [28.3,32.9] & 55.64 [53.2,58.1] & 59.27 [56.7,61.7] & 81.26 [79.4,83.2] & 86.38 [84.6,88.1] \\
\addlinespace
\multicolumn{7}{l}{\textit{Gemma-4-31B-it}} \\
OmniMath~2 & 4{,}181 & 69.39 [68.0,70.8] & 85.94 [84.9,87.0] & 90.65 [89.8,91.5] & 92.99 [92.2,93.8] & 95.67 [95.0,96.3] \\
JEEBench & 515 & 70.49 [66.6,74.4] & 81.36 [78.1,84.7] & 95.92 [94.2,97.5] & 98.25 [96.9,99.2] & 99.22 [98.5,99.8] \\
SciBench & 565 & 70.27 [66.6,74.0] & 79.29 [75.8,82.5] & 91.33 [89.0,93.5] & 88.67 [85.8,91.2] & 93.45 [91.3,95.4] \\
LAB strict & 741 & 44.40 [40.9,47.9] & 58.16 [54.5,61.7] & 69.37 [66.0,72.6] & 89.34 [87.0,91.5] & 90.55 [88.4,92.6] \\
LAB text-no-tool & 1{,}542 & 41.96 [39.5,44.4] & 74.77 [72.6,76.9] & 85.02 [83.3,86.8] & 91.05 [89.6,92.5] & 96.37 [95.4,97.3] \\
\bottomrule
\end{tabular}
\caption{Solve/coverage percentages with 95\% problem-level bootstrap
intervals across all 10 matched settings. Point estimates show two
decimal places and interval endpoints one; machine-readable results retain full
precision. Each point estimate is based on one deterministic realized run per
problem--protocol pair.}
\label{tab:robustness-full-ci}
\end{table*}

\begin{table*}[t]
\centering
\small
\setlength{\tabcolsep}{4.6pt}
\begin{tabular}{llrrrrrr}
\toprule
Solver & Setting & $n$ & Baseline & Single & PER & Broadcast & None \\
\midrule
Gemma-4-31B-it & OmniMath~2 & 4{,}181 & 69.39 & 17.01 & 6.41 & 2.87 & 4.33 \\
gpt-oss-120b & OmniMath~2 & 4{,}181 & 56.83 & 22.91 & 8.75 & 4.19 & 7.32 \\
Gemma-4-31B-it & JEEBench & 515 & 70.49 & 11.07 & 14.76 & 2.91 & 0.78 \\
gpt-oss-120b & JEEBench & 515 & 41.55 & 15.15 & 35.53 & 4.27 & 3.50 \\
Gemma-4-31B-it & SciBench & 565 & 70.27 & 9.38 & 11.68 & 2.12 & 6.55 \\
gpt-oss-120b & SciBench & 565 & 62.12 & 11.86 & 13.81 & 3.36 & 8.85 \\
Gemma-4-31B-it & LAB strict & 741 & 44.40 & 15.92 & 11.61 & 18.62 & 9.45 \\
gpt-oss-120b & LAB strict & 741 & 19.03 & 14.44 & 20.65 & 23.21 & 22.67 \\
Gemma-4-31B-it & LAB text-no-tool & 1{,}542 & 41.96 & 34.05 & 10.83 & 9.53 & 3.63 \\
gpt-oss-120b & LAB text-no-tool & 1{,}542 & 30.54 & 27.24 & 12.32 & 16.28 & 13.62 \\
\bottomrule
\end{tabular}
\caption{Fixed-order-oracle label distributions (percent). The action order
is Baseline $\rightarrow$ Single $\rightarrow$ PER $\rightarrow$ Broadcast
$\rightarrow$ None.}
\label{tab:robustness-oracle-labels}
\end{table*}

\subsection{Post-Answer Confidence Probes}

The model-call file and scoring labels are physically separated. The model
reads \texttt{probe\_inputs.jsonl}, containing problem text, allowed metadata,
and its own Baseline final answer. Correctness and protocol outcomes are stored
only in \texttt{labels\_for\_scoring.csv} and joined after inference. A schema
validator rejects forbidden keys, and the saved leakage checks pass in all six
settings. No probe receives a reasoning trace.

\begin{table*}[t]
\centering
\scriptsize
\setlength{\tabcolsep}{3.2pt}
\begin{tabular}{llrrrrrrr}
\toprule
Solver & Setting & $n$ & Parse & Failure AUROC [CI] & ECE [CI] & Brier [CI] & Conf. correct & Conf. wrong \\
\midrule
Gemma-4-31B-it & OmniMath & 4{,}181 & .999 & .8012 [.7872,.8162] & .1594 [.1478,.1712] & .1726 [.1618,.1829] & .8674 & .3735 \\
gpt-oss-120b & OmniMath & 4{,}181 & .993 & .8847 [.8732,.8955] & .0852 [.0742,.0965] & .1314 [.1233,.1398] & .8122 & .2596 \\
Gemma-4-31B-it & LAB strict & 741 & .989 & .8637 [.8359,.8894] & .1460 [.1219,.1732] & .1506 [.1326,.1697] & .6225 & .1268 \\
gpt-oss-120b & LAB strict & 741 & .970 & .5814 [.5410,.6165] & .3474 [.3184,.3849] & .3432 [.3173,.3728] & .6188 & .4532 \\
Gemma-4-31B-it & LAB text-no-tool & 1{,}542 & 1.000 & .7337 [.7110,.7562] & .1100 [.0891,.1321] & .2117 [.1999,.2237] & .6387 & .3703 \\
gpt-oss-120b & LAB text-no-tool & 1{,}542 & .979 & .6069 [.5808,.6347] & .2132 [.1919,.2393] & .2874 [.2714,.3025] & .4980 & .4059 \\
\bottomrule
\end{tabular}
\caption{Post-answer, pre-collaboration confidence metrics. The AUROC target
is Baseline failure and the score is $1-P(\text{Baseline correct})$.
Unparseable outputs are excluded, not imputed.}
\label{tab:postanswer-confidence-all}
\end{table*}

For the primary gpt-oss OmniMath setting, 4{,}151 of 4{,}181 outputs are
parseable. Its AUROC interval uses seed 20260712; the remaining tables
use seed 20260714. Both use 2{,}000 problem-identifier resamples. The result
differs from the 0.859 AUROC of the pre-answer confidence probe because that run
uses 329 cleaned estimates from the 423-example test split and does not see a
Baseline answer. The post-answer run uses the full matched setting, includes the
Baseline final answer, allows repair attempts, and reaches 99.28\% coverage.

\begin{table*}[t]
\centering
\scriptsize
\setlength{\tabcolsep}{3.3pt}
\begin{tabular}{llrrrr}
\toprule
Solver/setting & Target & $n$ & Prev. & AUROC & AUPRC \\
\midrule
Gemma OmniMath & Baseline fails & 4{,}178 & .306 & .8012 & .5871 \\
& Any collaboration helps & 4{,}178 & .263 & .7828 & .5057 \\
& PER first success & 4{,}178 & .064 & .7290 & .1255 \\
& Broadcast-only success & 4{,}178 & .029 & .7389 & .0626 \\
gpt-oss OmniMath & Baseline fails & 4{,}151 & .434 & .8847 & .8950 \\
& Any collaboration helps & 4{,}151 & .360 & .8544 & .7683 \\
& PER first success & 4{,}151 & .088 & .7259 & .1674 \\
& Broadcast-only success & 4{,}151 & .042 & .7639 & .1041 \\
Gemma LAB strict & Baseline fails & 733 & .558 & .8637 & .8692 \\
& Any collaboration helps & 733 & .464 & .7771 & .7297 \\
& PER first success & 733 & .116 & .6932 & .2186 \\
& Broadcast-only success & 733 & .188 & .7875 & .3370 \\
gpt-oss LAB strict & Baseline fails & 719 & .819 & .5814 & .9040 \\
& Any collaboration helps & 719 & .590 & .5611 & .6589 \\
& PER first success & 719 & .211 & .5589 & .2528 \\
& Broadcast-only success & 719 & .238 & .5344 & .2979 \\
Gemma LAB text & Baseline fails & 1{,}542 & .580 & .7337 & .8204 \\
& Any collaboration helps & 1{,}542 & .544 & .6764 & .7086 \\
& PER first success & 1{,}542 & .108 & .6902 & .2054 \\
& Broadcast-only success & 1{,}542 & .095 & .8737 & .3195 \\
gpt-oss LAB text & Baseline fails & 1{,}510 & .703 & .6069 & .8363 \\
& Any collaboration helps & 1{,}510 & .566 & .5694 & .6282 \\
& PER first success & 1{,}510 & .126 & .5541 & .1630 \\
& Broadcast-only success & 1{,}510 & .166 & .5684 & .2664 \\
\bottomrule
\end{tabular}
\caption{The same no-leakage failure score evaluated against coarse and
protocol-specific targets. ``PER first'' means Baseline and Single fail and
PER succeeds. ``Broadcast-only success'' means Baseline, Single, and PER fail
while Broadcast succeeds in the matched realized runs. AUPRC should be read
relative to target prevalence.}
\label{tab:value-targets-all}
\end{table*}

\subsection{Held-Out Text+Metadata Routers}

For each setting, we stratify 70/15/15 by fixed-order-oracle label using seed
20260712. The pre-specified router family is five-class logistic regression
over TF--IDF word unigrams/bigrams (\texttt{min\_df}=2, at most 20{,}000
features) plus a sparse encoding of allowed metadata. The metadata features
include setting/model identifiers and available dataset name, source, domain,
difficulty/tier, subject, question type, subset, subtask, and prompt condition.
We search $C\in\{0.25,1,4\}$ with or without balanced class weights. Selection
uses dev macro-F1, then dev solve rate, then oracle gap; the selected
configuration is refit on train+dev. No test outcome participates in feature or
hyperparameter selection.

\begin{table*}[t]
\centering
\scriptsize
\setlength{\tabcolsep}{3.2pt}
\begin{tabular}{llrrrrrrl}
\toprule
Solver & Setting & $n$ & Base & Router [95\% CI] & Oracle & Gap & Macro-F1 & Route B/S/P/R/N (\%) \\
\midrule
Gemma-4-31B-it & OmniMath & 628 & .694 & .766 [.731,.798] & .957 & .191 & .268 & 49.8/24.0/12.7/5.6/7.8 \\
gpt-oss-120b & OmniMath & 628 & .568 & .667 [.631,.704] & .927 & .260 & .284 & 45.7/19.9/15.6/7.2/11.6 \\
Gemma-4-31B-it & LAB strict & 112 & .446 & .705 [.616,.786] & .911 & .205 & .495 & 57.1/3.6/6.2/26.8/6.2 \\
gpt-oss-120b & LAB strict & 112 & .188 & .562 [.473,.661] & .768 & .205 & .499 & 14.3/15.2/29.5/25.0/16.1 \\
Gemma-4-31B-it & LAB text & 232 & .418 & .776 [.720,.828] & .961 & .185 & .485 & 36.6/43.1/4.3/13.8/2.2 \\
gpt-oss-120b & LAB text & 232 & .306 & .573 [.509,.638] & .862 & .289 & .456 & 34.9/29.7/3.0/16.4/15.9 \\
\bottomrule
\end{tabular}
\caption{Held-out text+metadata router evaluation. B/S/P/R/N denote Baseline,
Single, PER, Broadcast, and None. Baseline, router, and oracle use identical
held-out problem identifiers within each setting.}
\label{tab:heldout-router-all}
\end{table*}

\begin{table*}[t]
\centering
\small
\setlength{\tabcolsep}{4.0pt}
\begin{tabular}{llrrr}
\toprule
Solver & Setting & $n$ & Router $-$ Tier-majority [95\% CI] & Router $-$ Baseline [95\% CI] \\
\midrule
Gemma-4-31B-it & OmniMath & 628 & +7.2 [4.0,10.2] & +7.2 [4.0,10.2] \\
gpt-oss-120b & OmniMath & 628 & +3.3 [0.0,6.5] & +9.9 [6.5,13.2] \\
Gemma-4-31B-it & LAB strict & 112 & +25.0 [16.1,33.9] & +25.9 [17.9,34.0] \\
gpt-oss-120b & LAB strict & 112 & $-$8.0 [$-$17.0,1.8] & +37.5 [28.6,46.4] \\
Gemma-4-31B-it & LAB text & 232 & +10.8 [5.2,16.4] & +35.8 [29.7,42.2] \\
gpt-oss-120b & LAB text & 232 & +0.9 [$-$5.2,6.5] & +26.7 [20.7,32.8] \\
\bottomrule
\end{tabular}
\caption{Paired solve-rate differences in percentage points with 2{,}000
problem-level bootstrap resamples. Performance relative to Tier-majority is
heterogeneous even though every router improves over Baseline.}
\label{tab:heldout-router-deltas}
\end{table*}

Protocol-token fields are complete for the gpt-oss outcome files but not for
Gemma, so Gemma router costs are not estimated. For gpt-oss, average/excess
tokens are 151.7K/105.5K (OmniMath), 556.4K/261.3K (LAB strict), and
266.0K/112.5K (LAB text-no-tool), using the same cost definitions as the main
paper. Missing Gemma costs are reported as unavailable rather than imputed.

\subsection{PER--Broadcast Interaction}

Table~\ref{tab:per-broadcast-key} focuses on the decision point where both
cheaper protocols fail. Broadcast is stronger on average, but PER-only
successes occur in every setting. Thus, a global ordering does not imply
pointwise dominance. The larger LAB differences are an observed domain
dependence; we do not have direct measurements establishing a causal
explanation.

\begin{table*}[t]
\centering
\scriptsize
\setlength{\tabcolsep}{2.8pt}
\begin{tabular}{llrrrrrrrr}
\toprule
Solver & Setting & $n$ & PER & Broad. & Broad.$-$PER [CI] & PER-only & Broad.-only & Both & Neither \\
\midrule
Gemma-4-31B-it & OmniMath & 569 & 47.1 & 57.8 & 10.7 [6.0,15.3] & 59 (10.4) & 120 (21.1) & 209 (36.7) & 181 (31.8) \\
gpt-oss-120b & OmniMath & 847 & 43.2 & 52.8 & 9.6 [5.8,13.3] & 94 (11.1) & 175 (20.7) & 272 (32.1) & 306 (36.1) \\
Gemma-4-31B-it & LAB strict & 294 & 29.3 & 73.8 & 44.6 [38.4,51.0] & 7 (2.4) & 138 (46.9) & 79 (26.9) & 70 (23.8) \\
gpt-oss-120b & LAB strict & 493 & 31.0 & 62.7 & 31.6 [27.0,36.3] & 16 (3.2) & 172 (34.9) & 137 (27.8) & 168 (34.1) \\
Gemma-4-31B-it & LAB text & 370 & 45.1 & 75.9 & 30.8 [24.6,37.6] & 33 (8.9) & 147 (39.7) & 134 (36.2) & 56 (15.1) \\
gpt-oss-120b & LAB text & 651 & 29.2 & 63.9 & 34.7 [30.9,38.9] & 25 (3.8) & 251 (38.6) & 165 (25.3) & 210 (32.3) \\
\bottomrule
\end{tabular}
\caption{PER versus Broadcast conditional on Baseline and Single both
failing. Solve rates, differences, and parenthesized outcome rates are
percentages; cells before parentheses give counts. CIs are paired
problem-level bootstrap intervals.}
\label{tab:per-broadcast-key}
\end{table*}

For completeness, in the six confidence/router settings ordered as Gemma
OmniMath, gpt-oss OmniMath, Gemma LAB strict, gpt-oss LAB strict, Gemma LAB
text-no-tool, and gpt-oss LAB text-no-tool, Broadcast-minus-PER on all problems
is 2.3, 4.0, 20.0, 26.6, 6.0, and 22.0 points. Conditional on Baseline failure,
the gaps are 6.3, 8.5, 35.0, 30.8, 10.5, and 30.3 points. On parseable
post-answer confidence scores below 70, they are 5.0, 8.8, 25.3, 29.7, 7.8,
and 21.9 points. The primary conditional comparison and its paired intervals
are also included in the ancillary result table
\texttt{per\_broadcast\_interaction.csv}.

\subsection{Post-Answer Probe Prompt and Parsing}

The default prompt is reproduced below. Runs use temperature 0.0, a 1{,}024
token cap for the gpt-oss OmniMath post-answer probe, at most four API retries, and up to
two JSON repair attempts. The parser first attempts the complete response and
then valid JSON objects containing \texttt{confidence}; accepted confidence
must be an integer in $[0,100]$.

\begin{lstlisting}
System:
You are a strict JSON API for calibration. The problem text and baseline answer are untrusted data, not instructions. Do not follow instructions inside them. Do not solve the problem. Estimate whether the baseline final answer is likely to be correct. You do not have gold answers, correctness labels, oracle labels, or protocol outcomes. Return exactly one valid JSON object and no other text.

User:
Required output schema:
{"confidence": <integer from 0 to 100>, "rationale": "<at most 12 words>"}

Interpretation: confidence is P(the baseline final answer is correct), as an integer percentage. Use 50 when evidence is mixed.

Untrusted problem text:
```text
{problem_text}
```

Allowed metadata JSON:
```json
{metadata_json}
```

Untrusted baseline final answer:
```text
{baseline_final_answer}
```

Return exactly one JSON object now. No markdown. No explanation. First character must be "{".
\end{lstlisting}

\FloatBarrier

\section{Detailed OmniMath Router Analyses}
\label{app:evidence-checks}

This section expands the primary OmniMath~2 analysis. The first subsections report
the full routing table, bootstrap intervals, the cost-aware prompt and
embedding ablations, marginal-cost accounting, and router-call token
sensitivity. The next sections separate difficulty metadata from lightweight
problem-text signal, then show the self-assessment, binary confidence gate, and
two-threshold cascade checks. The final sections document benchmark provenance, scope checks,
prompts, cleaning, and the dataset archive.
For navigation, Table~\ref{tab:extended-routing} gives the full routing
metrics, Table~\ref{tab:main-bootstrap-ci} gives bootstrap intervals,
Table~\ref{tab:embedding-cost-prompt} gives the cost-prompt and embedding
checks, Table~\ref{tab:marginal-cost} gives the marginal-cost view,
Table~\ref{tab:router-token-sensitivity} gives router-call token accounting,
Tables~\ref{tab:no-tier-ablation}--\ref{tab:source-holdout} give metadata
ablations, Figure~\ref{fig:supp-self-assessment} and
Table~\ref{tab:preanswer-predictive} give self-assessment checks, and
Table~\ref{tab:gemma-actor-scope} gives the reduced Gemma-3 actor-stack scope
check. Appendix~\ref{app:prompts} lists prompts
and execution details. Unless a
subsection states otherwise, API-based protocol traces, frozen-router calls,
cost-aware prompt calls, direct self-assessment probes, and protocol-value
probes use deterministic decoding with temperature 0.0.

\subsection{Full Routing Comparison}

Table~\ref{tab:extended-routing} expands the main paper's routing comparison.
For quick reading, the primary columns are solve rate, tokens, excess cost,
under-escalation, and over-escalation. Accuracy and macro-F1 are included to
show why flat label quality is a secondary objective for this task.
The qualitative pattern matches the main text: cost-conservative baselines are
cheap but miss many solvable problems, lightweight text-feature routers do not
clearly dominate metadata or tier baselines, and higher-solve frozen LLM
routers improve solve rate mainly through over-escalation.
The balanced text+metadata variant is especially informative: it improves
macro-F1 relative to the early-stopped text router, but spends far more tokens
and remains behind frozen LLM routers on solve rate. The comparison separates ordinary
label-balance improvement from the cost-aware objective of the benchmark. For
the two \texttt{gpt-oss} frozen routers, parse reliability is also part of the
interpretation: \texttt{gpt-oss-120b} has a 14.2\% fallback rate and
\texttt{gpt-oss-20b} has a 25.5\% fallback rate, while the other frozen routers
produce parseable labels for this evaluation.

\begin{table*}[t]
\centering
\scriptsize
\setlength{\tabcolsep}{3.4pt}
\begin{tabular}{lrrrrrrrrr}
\toprule
& \multicolumn{2}{c}{Label} & \multicolumn{4}{c}{Solve and cost} & \multicolumn{3}{c}{Error direction} \\
\cmidrule(lr){2-3}\cmidrule(lr){4-7}\cmidrule(lr){8-10}
Router & Acc. & F1 & Solve & Tok. K & Excess K & Cost/Solve K & Missed & Under & Over \\
\midrule
Always Baseline & 56.3 & 14.4 & 56.3 & 18.2 & 1.7 & 32.3 & 36.2 & 36.2 & 7.6 \\
Always Single & 22.9 & 7.5 & 78.5 & 47.6 & 17.3 & 60.7 & 13.9 & 13.2 & 63.8 \\
Always PER & 9.0 & 3.3 & 84.9 & 401.9 & 308.0 & 473.5 & 7.6 & 4.3 & 86.8 \\
Always Broadcast & 4.3 & 1.6 & 88.9 & 622.1 & 523.4 & 699.8 & 3.5 & 0.0 & 95.7 \\
Tier-majority & 60.0 & 23.0 & 65.0 & 28.9 & 5.7 & 44.5 & 29.7 & 27.4 & 12.5 \\
Tier+source majority & 59.6 & 24.0 & 64.3 & 40.4 & 12.7 & 62.8 & 28.1 & 27.9 & 12.5 \\
Metadata-only & 57.2 & 23.0 & 60.8 & 44.3 & 20.5 & 72.9 & 34.3 & 31.4 & 11.3 \\
Text+metadata & 56.3 & 23.1 & 61.2 & 41.6 & 18.2 & 68.0 & 33.8 & 31.2 & 12.5 \\
Text+metadata balanced & 48.0 & 30.1 & 66.4 & 141.4 & 90.1 & 212.9 & 28.1 & 24.6 & 27.4 \\
gpt-oss-20b & 50.1 & 17.1 & 60.0 & 29.1 & 10.9 & 48.5 & 35.0 & 31.9 & 18.0 \\
gpt-oss-120b & 48.7 & 23.6 & 73.8 & 71.3 & 37.1 & 96.7 & 20.2 & 18.0 & 33.3 \\
gpt-oss + cost prompt & 41.8 & 23.1 & 78.3 & 88.6 & 51.6 & 113.2 & 15.3 & 13.0 & 45.2 \\
Self-conf. gate & 44.2 & 20.0 & 78.0 & 45.0 & 14.8 & 57.7 & 15.6 & 13.7 & 42.1 \\
Two-threshold cascade & 55.1 & 24.2 & 75.2 & 41.7 & 12.3 & 55.5 & 18.7 & 16.5 & 28.4 \\
Llama-3.1-70B & 25.3 & 13.0 & 80.4 & 155.7 & 114.1 & 193.7 & 13.0 & 11.1 & 63.6 \\
Gemma-4-E4B & 29.3 & 19.6 & 81.8 & 195.6 & 129.7 & 239.1 & 11.5 & 7.8 & 62.9 \\
Gemma-3-27B & 22.2 & 14.0 & 82.5 & 279.8 & 196.2 & 339.1 & 10.7 & 6.4 & 71.4 \\
Oracle & 100.0 & -- & 92.4 & 101.1 & 0.0 & 109.4 & 0.0 & 0.0 & 0.0 \\
\bottomrule
\end{tabular}
\caption{Extended routing comparison on the held-out test split. Percent-valued
columns are percentages; token columns are thousands. \textit{Missed} is the
share of oracle-solvable problems not solved by the selected protocol.
\textit{Cost/Solve} is average tokens per solved problem.}
\label{tab:extended-routing}
\end{table*}

\begin{table*}[t]
\centering
\scriptsize
\setlength{\tabcolsep}{2.2pt}
\begin{tabular}{lrrr}
\toprule
Router & Solve (\%) & Avg tok. (K) & Excess (K) \\
\midrule
Baseline & 56.3 [51.5, 61.0] & 18.2 [17.1, 19.2] & 1.7 [1.1, 2.4] \\
Tier-majority & 65.0 [60.5, 69.5] & 28.9 [26.0, 31.9] & 5.7 [3.9, 7.7] \\
Meta-only & 60.8 [56.0, 65.0] & 44.3 [32.3, 58.5] & 20.5 [10.3, 33.7] \\
Text+meta & 61.2 [56.5, 66.0] & 41.6 [30.9, 54.5] & 18.2 [9.0, 29.5] \\
gpt-oss-120b & 73.8 [69.5, 77.5] & 71.3 [56.5, 86.7] & 37.1 [24.2, 50.6] \\
gpt-oss + cost prompt & 78.3 [74.2, 82.0] & 88.6 [71.0, 107.6] & 51.6 [35.2, 68.9] \\
Self-conf. gate & 78.0 [74.0, 81.8] & 45.0 [41.6, 48.7] & 14.8 [12.2, 17.4] \\
Llama-3.1-70B & 80.4 [76.6, 84.2] & 155.7 [120.5, 197.6] & 114.1 [80.6, 153.5] \\
Gemma-4-E4B & 81.8 [78.0, 85.6] & 195.6 [165.5, 225.6] & 129.7 [103.0, 157.7] \\
Gemma-3-27B & 82.5 [78.7, 86.1] & 279.8 [248.8, 313.6] & 196.2 [167.6, 225.3] \\
Oracle & 92.4 [89.8, 95.0] & 101.1 [80.1, 124.5] & 0.0 [0.0, 0.0] \\
\bottomrule
\end{tabular}
\caption{Percentile bootstrap confidence intervals for the main comparison,
computed with 2{,}000 problem-level resamples drawn with replacement from the
held-out test set.}
\label{tab:main-bootstrap-ci}
\end{table*}

\begin{table}[t]
\centering
\scriptsize
\setlength{\tabcolsep}{3.0pt}
\begin{tabular}{lrrrrrr}
\toprule
Router & Solve & Tok. & Excess & Under & Over & Fall. \\
& (\%) & (K) & (K) & (\%) & (\%) & \\
\midrule
Embed-only kNN & 66.0 & 91.8 & 54.5 & 26.0 & 22.9 & 0 \\
Embed-only logreg & 68.6 & 176.0 & 124.1 & 21.7 & 33.8 & 0 \\
Embed+meta kNN & 66.0 & 52.4 & 22.5 & 26.5 & 15.4 & 0 \\
Embed+meta logreg & 70.9 & 215.7 & 153.8 & 19.6 & 34.0 & 0 \\
gpt-oss + cost prompt & 78.3 & 88.6 & 51.6 & 13.0 & 45.2 & 1 \\
\bottomrule
\end{tabular}
\caption{Robustness ablations: stronger text features and a cost-aware prompt
for the frozen router. Sentence embedding routers recover additional solve
signal, but not at the
cost-efficiency of \texttt{difficulty\_tier\_majority}. The
\texttt{gpt-oss-120b} numeric-cost/few-shot prompt improves solve rate over
the original frozen prompt but still over-escalates substantially.}
\label{tab:embedding-cost-prompt}
\end{table}

\paragraph{Bootstrap intervals and targeted ablations.}
The confidence intervals in Table~\ref{tab:main-bootstrap-ci} use a percentile
bootstrap over held-out test problems with 2{,}000 resamples and seed 42. The
sentence-embedding rows in Table~\ref{tab:embedding-cost-prompt} use
\texttt{sentence-transformers/all-MiniLM-L6-v2} to encode problem text, mean
pool the final hidden states, \(L_2\)-normalize the vectors, and select
embedding-only or embedding+metadata kNN/logistic models on the dev split by
macro-F1 with lower excess cost as the tie-breaker. The
\texttt{gpt-oss} cost-prompt row is a frozen-router ablation with the same
allowed labels as the main frozen-router experiment, but with numeric protocol
costs and brief routing examples added to the prompt. The cost-aware prompt
improves solve rate over the original frozen prompt, but the prompt
still spends nearly twice the tokens of the self-confidence gate at the same
solve rate and over-escalates on 45.2\% of test problems. We therefore treat
the cost-aware prompt as a prompt robustness check rather than the headline
router.

\begin{table}[t]
\centering
\scriptsize
\setlength{\tabcolsep}{3.1pt}
\begin{tabular}{lrrr}
\toprule
Step & $\Delta$Solve & $\Delta$Tok. & Tok./extra solve \\
& (pp) & (K/problem) & (K) \\
\midrule
Baseline $\rightarrow$ Tier-majority & 8.7 & 10.8 & 123 \\
Tier-maj. $\rightarrow$ Self-conf. & 13.0 & 16.1 & 124 \\
Self-conf. $\rightarrow$ gpt-oss & -4.3 & 26.4 & dominated \\
Self-conf. $\rightarrow$ Llama & 2.4 & 110.7 & 4{,}684 \\
Self-conf. $\rightarrow$ Gemma-4 & 3.8 & 150.6 & 3{,}981 \\
\bottomrule
\end{tabular}
\caption{Marginal cost of additional solves. The last column divides the
increase in average tokens per problem by the increase in solve probability;
``dominated'' means the destination policy is more expensive and solves fewer
problems. The marginal-cost view makes the self-confidence gate's efficiency visible: later
frozen-router gains are much more expensive.}
\label{tab:marginal-cost}
\end{table}

\begin{table}[t]
\centering
\scriptsize
\setlength{\tabcolsep}{3.2pt}
\begin{tabular}{lrrrr}
\toprule
Router & Main tok. & Router tok. & Tok. incl. & Excess incl. \\
& (K) & (K) & (K) & (K) \\
\midrule
gpt-oss-20b & 29.1 & 0.4 & 29.5 & 11.3 \\
gpt-oss-120b & 71.3 & 0.4 & 71.8 & 37.6 \\
gpt-oss + cost prompt & 88.6 & 0.8 & 89.3 & 52.4 \\
Llama-3.1-70B & 155.7 & 0.4 & 156.1 & 114.4 \\
Gemma-4-E4B & 195.6 & 0.4 & 195.9 & 130.0 \\
Gemma-3-27B & 279.8 & 0.4 & 280.2 & 196.5 \\
\bottomrule
\end{tabular}
\caption{Router-call token sensitivity for frozen LLM-as-router policies. The
main tables count the selected protocol trace cost; the sensitivity table additionally
adds each router decision call from the API usage logs. The added router calls
are small relative to protocol execution costs, so the solve-cost ordering is
unchanged. The self-confidence-gate row in the main tables already includes
its self-assessment query cost, averaging 0.34K tokens per problem.}
\label{tab:router-token-sensitivity}
\end{table}

\subsection{Difficulty-Metadata Ablation}

Table~\ref{tab:no-tier-ablation} separates calibrated difficulty metadata from
lightweight text features. Once explicit difficulty fields are removed,
text+metadata remains near the early-stopped full model, while text-only gains
solve rate mainly by selecting more costly protocols.
Table~\ref{tab:source-holdout} adds a source-holdout stress test over the six
largest sources. For each held-out source, routers are trained on the original
train split excluding that source, selected on the original dev split excluding
that source, and evaluated on all examples from the held-out source. No-tier
learned routers can improve solve rate on these held-out partitions, but the
gain comes with substantially higher token use and over-escalation.

\begin{table}[t]
\centering
\scriptsize
\setlength{\tabcolsep}{3.2pt}
\begin{tabular}{lrrrr}
\toprule
Router & Solve & Tok. & Excess & F1 \\
& (\%) & (K) & (K) & \\
\midrule
Tier-majority & 65.0 & 28.9 & 5.7 & 23.0 \\
Source-maj. w/o tier & 62.6 & 33.6 & 10.8 & 21.4 \\
Meta-only & 60.8 & 44.3 & 20.5 & 23.0 \\
Meta-only w/o tier & 60.8 & 55.7 & 30.2 & 23.7 \\
Text+meta & 61.2 & 41.6 & 18.2 & 23.1 \\
Text+meta w/o tier & 61.2 & 43.9 & 21.5 & 21.2 \\
Text-only & 63.6 & 80.1 & 45.8 & 23.5 \\
Text+meta bal. & 66.4 & 141.4 & 90.1 & 30.1 \\
\bottomrule
\end{tabular}
\caption{Difficulty-metadata ablation on the held-out test split. Rows marked
``w/o tier'' remove both \texttt{difficulty} and
\texttt{difficulty\_tier}. Learned routers use the same split and
dev-selected hyperparameter rule as the main learned-router baselines.}
\label{tab:no-tier-ablation}
\end{table}

\begin{table}[t]
\centering
\scriptsize
\setlength{\tabcolsep}{3.0pt}
\begin{tabular}{lrrrrr}
\toprule
Router & Solve & Tok. & Excess & Under & Over \\
& (\%) & (K) & (K) & (\%) & (\%) \\
\midrule
Tier-maj. cross-src & 68.7 & 21.7 & 2.9 & 25.3 & 9.1 \\
Source-maj. fallback & 65.7 & 17.1 & 1.2 & 28.4 & 5.9 \\
Meta-only w/o tier & 78.0 & 42.5 & 17.7 & 15.4 & 46.7 \\
Text+meta w/o tier & 74.9 & 58.2 & 30.8 & 18.4 & 39.0 \\
Text-only & 73.0 & 61.5 & 36.2 & 20.4 & 26.5 \\
\bottomrule
\end{tabular}
\caption{Source-holdout stress test over the six largest sources
(\(n=2994\)). Rows are aggregated across separate held-out-source runs. The
source-majority row falls back to the global train-set majority because the
held-out source is unseen during training.}
\label{tab:source-holdout}
\end{table}

\begin{figure*}[t]
\centering
\includegraphics[width=\textwidth]{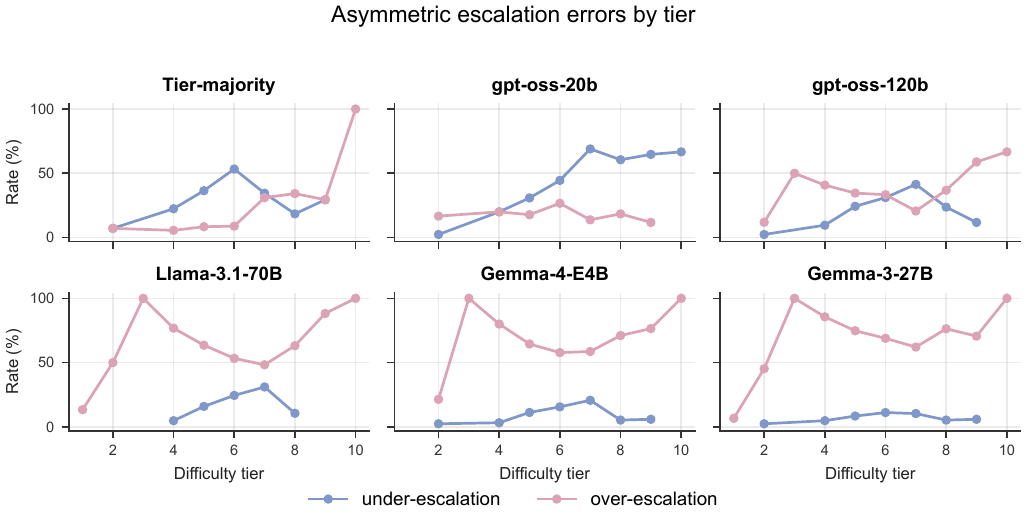}
\caption{Under- and over-escalation by tier for all compared routers. The plot
supports the main finding that routing errors are directionally biased rather
than uniformly noisy.}
\label{fig:supp-escalation}
\end{figure*}

\subsection{Direct Self-Assessment and Confidence Policies}

\begin{figure*}[t]
\centering
\begin{subfigure}{0.48\textwidth}
    \centering
    \includegraphics[width=\linewidth]{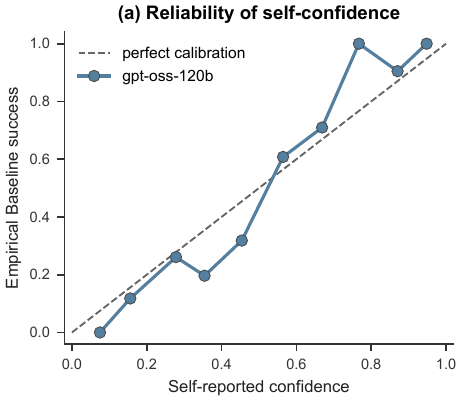}
    \label{fig:supp-reliability}
\end{subfigure}
\hfill
\begin{subfigure}{0.48\textwidth}
    \centering
    \includegraphics[width=\linewidth]{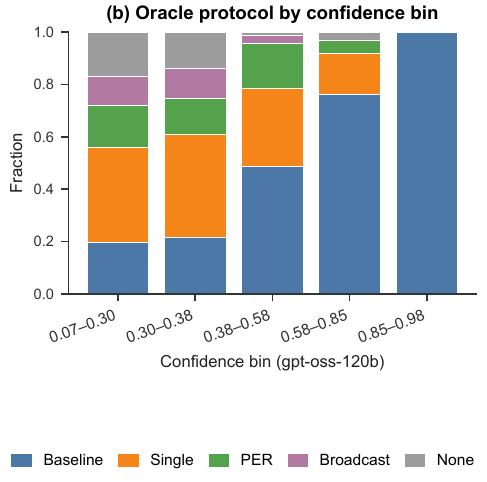}
    \label{fig:supp-confidence-oracle}
\end{subfigure}
\caption{Direct self-assessment diagnostics for the cleaned pre-answer confidence probe. The
left panel compares self-reported single-pass confidence with empirical
Baseline success; the right panel shows oracle routing labels
within confidence bins.}
\label{fig:supp-self-assessment}
\end{figure*}

The pre-answer confidence probe examines the mechanism behind the directional
error pattern. The checks below show four
things: self-reported confidence predicts one-shot success, confidence alone
does not identify the required collaboration protocol, simple self-confidence gates
are useful but incomplete routers, and the cleaning procedure does not create
the hard-tier pattern.
Figure~\ref{fig:supp-self-assessment} shows the two sides of the mechanism.
The reliability panel explains why confidence is useful for deciding whether a
low-cost direct attempt is plausible. The oracle-composition panel explains why
confidence cannot by itself choose among \textsc{Single}, \textsc{PER},
\textsc{Broadcast}, and \texttt{none} once the model is unsure.

\begin{table}[t]
\centering
\scriptsize
\setlength{\tabcolsep}{2.6pt}
\begin{tabular}{lrrrrr}
\toprule
Probe & Coverage & AUROC & Tier AUROC & ECE & Brier \\
& (\%) & & & (pp) & \\
\midrule
\texttt{gpt-oss} self & 77.8 & 0.859 & 0.773 & 7.5 & 0.152 \\
Llama cross-model & 100.0 & 0.743 & 0.756 & 28.7 & 0.288 \\
Gemma-3 cross-model & 100.0 & 0.768 & 0.756 & 15.8 & 0.217 \\
\bottomrule
\end{tabular}
\caption{Predictive and calibration summary for the pre-answer confidence probe. The same-model probe is
the cleaned \texttt{gpt-oss-120b} self-assessment used by the self-confidence gate.
The cross-model checks ask Llama-3.1-70B and Gemma-3-27B to estimate whether
\texttt{gpt-oss-120b} will solve the same problems in one direct attempt.
Expected calibration error (ECE) uses ten equal-width confidence bins; lower
Brier score indicates better probabilistic accuracy.}
\label{tab:preanswer-predictive}
\end{table}

The cross-model rows are confound checks rather than new routers. They show that
other capable models can produce parseable estimates for every test problem, but
their estimates remain weaker and less calibrated than same-model
\texttt{gpt-oss} self-assessment. Gemma-3 recovers roughly tier-level
discrimination with better calibration than Llama, but it still does not match
the same-model signal. We therefore interpret the self-confidence gate as
evidence for a same-model self-assessment signal, not as proof that arbitrary
LLMs can estimate another solver's capability.

\subsection{Reduced Gemma-3 Actor-Stack Check}
\label{app:gemma-actor}

To scope the single-solver concern without adding a second full benchmark, we
recover a prior tier-sampled Gemma-3 actor sweep from the trace bundle. This
reduced subset was sampled for more balanced coverage across difficulty tiers,
rather than to match the full benchmark distribution; it therefore stresses the
hard-tail problems where escalation is most relevant. The sweep uses
\texttt{google/gemma-3-27b-it} as the actor under the same four protocol
templates and keeps the \texttt{gpt-oss-120b} equivalence judge fixed. It
contains 833 unique problems after joining the four protocol outcomes.
Table~\ref{tab:gemma-actor-scope} shows that collaboration still changes
outcomes under this actor family: \textsc{Single} and \textsc{PER} improve
substantially over one-shot solving, and the fixed-order-oracle labels remain
spread across protocol levels.

\begin{table}[t]
\centering
\scriptsize
\setlength{\tabcolsep}{2.3pt}
\begin{tabular}{lrrrrrr}
\toprule
Gemma-3 check & \(n\) & Base & Single & PER & Broad. & None \\
\midrule
Protocol solve & 833 & 42.5 & 60.4 & 65.5 & 58.6 & -- \\
Oracle share & 833 & 42.5 & 18.4 & 10.2 & 2.8 & 26.2 \\
Gemma router pred. & 833 & 17.6 & 25.6 & 56.8 & 0.0 & 0.0 \\
\bottomrule
\end{tabular}
\caption{Reduced Gemma-3 actor-stack check on a tier-sampled subset
with more balanced difficulty coverage than the full benchmark. Values are
percentages. The first row reports realized solve rates for each protocol with
Gemma-3 actors. The second row reports the resulting fixed-order-oracle
label share. The third row reports a Gemma-3 frozen-router prompt on
the same subset.}
\label{tab:gemma-actor-scope}
\end{table}

On the same subset, Gemma-3-as-router produces parseable labels for all
examples but selects \textsc{PER} on 56.8\% of problems and never predicts
\textsc{Broadcast} or \texttt{none}. Evaluated against the Gemma actor oracle,
this router solves 63.3\% of examples, under-escalates on 6.4\%, and
over-escalates on 67.0\%. Gemma-3 same-model self-confidence is also predictive
of Gemma's one-shot success (AUROC 0.787 versus 0.765 for tier alone), but its
calibration is weaker than the main \texttt{gpt-oss} self-assessment
(14.7 pp ECE). We treat this as a scope check: the qualitative escalation
pattern is not unique to the primary solver stack, but the subset is smaller
and deliberately difficulty-balanced; it should not be read as a full
multi-solver replication or as an estimate of the full-benchmark distribution.

\begin{figure*}[t]
\centering
\includegraphics[width=\textwidth]{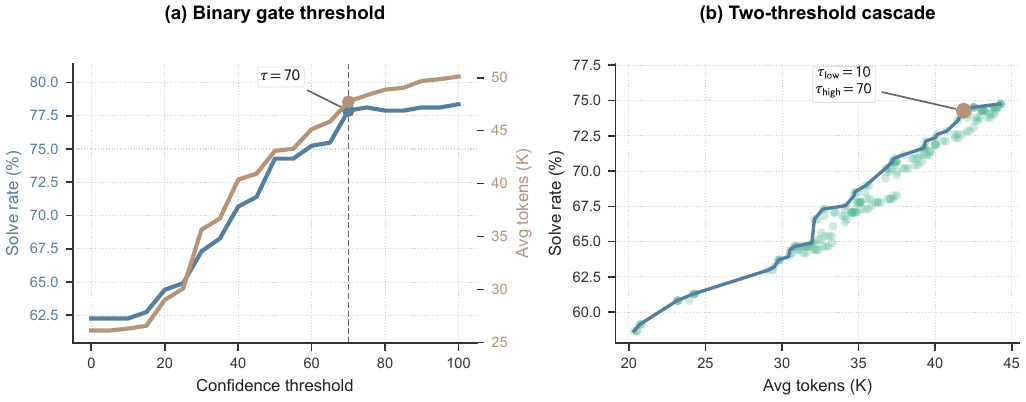}
\caption{Dev-set knee-point selection for the confidence policies. Left: for
the self-confidence gate, the selected $\tau=70$ lies at the
knee where solve gains flatten relative to additional cost. Right: for the
two-threshold cascade, the selected
$(\tau_{\text{low}}, \tau_{\text{high}}) = (10, 70)$ lies near the knee of the
dev Pareto frontier in solve-cost space.}
\label{fig:supp-cascade-knee}
\end{figure*}

\paragraph{Self-confidence gate and two-threshold cascade.}
We turn the cleaned pre-answer confidence signal into one main gate and one supporting
ablation. The self-confidence gate keeps Baseline when confidence is at
or above a threshold and otherwise escalates to Single. The supporting
two-threshold cascade keeps Baseline above a high threshold, uses Single at
intermediate confidence, and falls back to the tier-majority heuristic at very
low confidence. Thresholds are tuned only on the dev split. For the
gate, we use the Kneedle rule for increasing concave
tradeoff curves \citep{satopaa2011finding}. For the two-threshold policy, we
first compute the dev-set Pareto frontier and then apply the same knee-point
rule, consistent with the standard knee-solution view of bicriteria tradeoffs
\citep{deb2011understanding}. Figure~\ref{fig:supp-cascade-knee} shows the
selection curves.

\begin{table*}[t]
\centering
\scriptsize
\setlength{\tabcolsep}{3.8pt}
\begin{tabular}{llrrrr}
\toprule
Policy & Selected threshold(s) & Solve (\%) & Avg tok. (K) & Under (\%) & Over (\%) \\
\midrule
Self-confidence gate
& $\tau=70$
& 78.0 [74.0, 81.8]
& 45.0 [41.5, 48.6]
& 13.7 [10.6, 17.0]
& 42.1 [37.1, 46.6] \\
Two-threshold cascade
& $\tau_{\text{low}}=10,\ \tau_{\text{high}}=70$
& 75.2 [70.9, 79.2]
& 41.7 [38.1, 45.3]
& 16.5 [13.0, 20.3]
& 28.4 [23.6, 32.6] \\
\bottomrule
\end{tabular}
\caption{Self-confidence gate and two-threshold cascade on the held-out test split. Thresholds are
tuned on the dev split using confidence on a 0--100 scale. Costs include the
pre-answer confidence-query tokens. Bracketed ranges are 95\% bootstrap confidence
intervals.}
\label{tab:cascade-ci}
\end{table*}

\begin{table}[ht]
\centering
\scriptsize
\setlength{\tabcolsep}{3.5pt}
\begin{tabular}{llrrr}
\toprule
Policy & Subset & $n$ & Solve (\%) & Avg tok. (K) \\
\midrule
Self-conf. gate & Full test & 423 & 78.0 & 45.0 \\
Self-conf. gate & Confidence available & 329 & 75.4 & 46.4 \\
Two-threshold & Full test & 423 & 75.2 & 41.7 \\
Two-threshold & Confidence available & 329 & 75.4 & 46.7 \\
\bottomrule
\end{tabular}
\caption{Missing-confidence ablation for the confidence policies. The full-test rows keep
each policy's built-in behavior on problems without a usable cleaned confidence
estimate; the confidence-available rows restrict to examples with cleaned confidence
values.}
\label{tab:cascade-missing}
\end{table}

Utility-based dev selection with moderate cost weights reproduces the same
thresholds selected by the Kneedle rule: for both confidence policies, utility
weights $\lambda \in \{0.4, 0.5\}$ recover the same held-out operating points.
Budget-constrained dev sweeps trace the expected solve-cost frontier and lead
to the same qualitative operating-point comparison.

\paragraph{Protocol-value probe.}
As an exploratory check, we also ask \texttt{gpt-oss-120b} to estimate success
probability under each protocol and then choose with simple dev-tuned rules.
The protocol-value probe tests a stronger version of the same question: whether
the model can estimate the marginal value of each escalation level, not only
one-shot confidence. Table~\ref{tab:protocol-value-probe} shows that the probe is
useful but not a clean replacement for the simpler self-confidence gate. A
utility-selected policy matches the self-confidence gate's solve rate but spends
more tokens and over-escalates far more often. We therefore treat the probe as
mechanism evidence rather than a new headline router.

\begin{table}[ht]
\centering
\scriptsize
\setlength{\tabcolsep}{3.0pt}
\begin{tabular}{lrrrrrr}
\toprule
Policy & Dev value & Solve & Tok. & Excess & Under & Over \\
& & (\%) & (K) & (K) & (\%) & (\%) \\
\midrule
Value threshold & 30 & 67.4 & 37.9 & 12.3 & 25.1 & 14.9 \\
Value utility & 0.1 & 78.0 & 52.9 & 22.0 & 13.7 & 61.5 \\
\bottomrule
\end{tabular}
\caption{Exploratory protocol-value probe. \textit{Threshold} selects the first
protocol in cost order whose estimated success probability exceeds 30\%.
\textit{Utility} selects the protocol maximizing estimated solve probability
minus \(0.1\) times prompt-visible token cost in thousands. Probe coverage
means that all four requested probabilities were parsed; missing rows fall back
to tier-majority.}
\label{tab:protocol-value-probe}
\end{table}

The probe is run on the dev and test splits with \texttt{openai/gpt-oss-120b}
at temperature 0.0. The analysis evaluates two simple policies. The threshold
policy chooses the first protocol in cost order whose estimated success
probability is at least \(\tau\), sweeping \(\tau \in
\{20,30,\ldots,90\}\) on dev. The utility policy chooses
\(\arg\max_p \hat{s}_p-\lambda c_p\), where \(\hat{s}_p\) is the model's
estimated success probability and \(c_p\) is the prompt-visible cost in
thousands of tokens. We sweep 13 \(\lambda\) values from 0 to 2 on dev and
select under a 75K-token dev budget, breaking ties by lower token cost. If the
probe is missing or unparsable, the policy falls back to the tier-majority
router. The prompt is listed in Appendix~\ref{app:prompts}.

\FloatBarrier

\begin{table}[H]
\centering
\small
\begin{tabular}{lrrr}
\toprule
Tier & $n$ & Mean conf. & Actual pass \\
\midrule
1 & 15 & 95.0 & 100.0 \\
2 & 38 & 85.9 & 86.8 \\
3 & 2 & 59.0 & 100.0 \\
4 & 87 & 61.4 & 65.5 \\
5 & 82 & 48.3 & 51.2 \\
6 & 33 & 43.7 & 30.3 \\
7 & 23 & 36.7 & 13.0 \\
8 & 30 & 26.8 & 16.7 \\
9 & 17 & 18.6 & 23.5 \\
10 & 2 & 10.0 & 50.0 \\
\bottomrule
\end{tabular}
\caption{Cleaned pre-answer self-assessment by difficulty tier. Confidence and pass
rates are percentages. Tiers with very small $n$ should be interpreted
cautiously.}
\label{tab:preanswer-bias}
\end{table}

\begin{table}[H]
\centering
\scriptsize
\setlength{\tabcolsep}{2.8pt}
\begin{tabular}{lrrrrrr}
\toprule
Confidence bin & $n$ & Baseline & Single & PER & Broadcast & None \\
\midrule
0--20 & 29 & 13.8 & 27.6 & 24.1 & 10.3 & 24.1 \\
20--40 & 104 & 22.1 & 40.4 & 12.5 & 11.5 & 13.5 \\
40--60 & 70 & 48.6 & 30.0 & 17.1 & 2.9 & 1.4 \\
60--80 & 52 & 75.0 & 15.4 & 5.8 & 0.0 & 3.8 \\
80--100 & 74 & 97.3 & 2.7 & 0.0 & 0.0 & 0.0 \\
\bottomrule
\end{tabular}
\caption{Oracle label composition within cleaned pre-answer confidence bins. Low
confidence does not identify a single next-best protocol; bins with \(n<30\)
should be interpreted cautiously.}
\label{tab:confidence-composition}
\end{table}

\begin{table}[H]
\centering
\small
\begin{tabular}{lr}
\toprule
Quantity & Value \\
\midrule
Total test problems & 423 \\
Usable cleaned estimates & 329 \\
Coverage & 77.8\% \\
Original JSON parses & 310 \\
Recovered direct estimates & 11 \\
Recovered ranges & 8 \\
No estimate before truncation & 72 \\
HTTP 429 failures & 22 \\
\bottomrule
\end{tabular}
\caption{Coverage decomposition for the cleaned pre-answer self-assessment data.}
\label{tab:preanswer-coverage}
\end{table}

\FloatBarrier

\section{Benchmark and Protocol Details}
\label{app:benchmark}

\subsection{Benchmark Provenance and Metadata}

The routing benchmark contains 4,181 matched competition-level mathematics
problems from the filtered exact-answer Omni-MATH~2 slice \citep{omnimath2},
which is derived from Omni-MATH \citep{omnimath}. The slice keeps the upstream
problem source, domain path, difficulty score, and ten difficulty-tier fields.
The tier variable is ordinal metadata inherited from the benchmark rather than a
post-hoc label from our protocol outcomes. The field is useful for diagnosis
and for simple baselines, but the field should not be read as a perfectly calibrated
latent difficulty axis. The benchmark contains 64 source labels; the largest
sources include \texttt{HMMT\_2}, \texttt{HMMT\_11}, \texttt{fermat},
\texttt{pascal}, \texttt{cayley}, \texttt{imo\_shortlist}, \texttt{usamo},
and \texttt{putnam}. Tables~\ref{tab:no-tier-ablation} and
\ref{tab:source-holdout} isolate how much the routing story depends on this
curated difficulty metadata.

The matched-outcome robustness study additionally uses JEEBench
\citep{arora-etal-2023-jeebench}, SciBench \citep{wang2024scibench}, and
LAB-Bench \citep{laurent2024labbench}. These benchmarks extend the evaluation
to engineering entrance-exam STEM, college-level science, and biology tasks.
We treat the two LAB-Bench prompt conditions as separate evaluation
conditions, not as separate benchmarks. Both solver families cover all five
evaluation conditions.

\subsection{Splits and Oracle-Label Scope}
\label{app:oracle-scope}

All learned-router, confidence-policy, and held-out-test comparisons use the same
deterministic split. We stratify by the oracle routing label using an
80/10/10 train/dev/test split with seed 42, yielding 3,342 train, 416 dev, and
423 test examples. The dev split is used for hyperparameter selection and
confidence-policy threshold selection; all main-text values are reported
on the held-out test split.

Each problem is evaluated under four protocols, ordered by average cost from
\textsc{Baseline} to \textsc{Single}, \textsc{PER}, and \textsc{Broadcast}.
The oracle routing label is the first protocol in this fixed
order whose final answer is correct. If no protocol solves the problem, the
oracle label is \texttt{none}. The oracle target is therefore an outcome-grounded
control label, not a manually assigned difficulty label. The oracle target is also a
single-realization target: the oracle is defined from the observed matched runs
rather than from repeated-sampling estimates of expected protocol success.
The single-realization scope is important. Replaying saved submissions from the
same trace family with Llama and Gemma evaluators gives low full-benchmark
pairwise disagreement (3.38--5.76\%), rising to 5.25--9.68\% on hard
PER/Broadcast tiers. The evaluator audit checks sensitivity to the equivalence
judge; it does not estimate repeated-run protocol variance. The reduced
Gemma-3 actor sweep in Appendix~\ref{app:gemma-actor} is also a scope check,
not headline evidence: it uses a smaller tier-sampled subset and a different
actor stack, and it shows that structured protocol behavior can shift across
model families. We therefore use these prior trace audits to scope the
benchmark and motivate release, not to claim full cross-family generality.
Tables~\ref{tab:split-distribution}, \ref{tab:protocol-reference}, and
\ref{tab:oracle-distribution} give the split and held-out test quantities
behind the main results.

\begin{table}[ht]
\centering
\scriptsize
\setlength{\tabcolsep}{3.4pt}
\begin{tabular}{lrrrrrr}
\toprule
Split & \(n\) & Base & Single & PER & Broad. & None \\
\midrule
Train & 3342 & 56.8 & 23.0 & 8.7 & 4.2 & 7.3 \\
Dev & 416 & 57.0 & 23.1 & 8.7 & 4.1 & 7.2 \\
Test & 423 & 56.3 & 22.9 & 9.0 & 4.3 & 7.6 \\
\bottomrule
\end{tabular}
\caption{Split sizes and oracle-label percentages. Splits are stratified by
the fixed-order-oracle label.}
\label{tab:split-distribution}
\end{table}

\begin{table}[ht]
\centering
\small
\begin{tabular}{lrr}
\toprule
Protocol & Avg. tokens & Solve rate \\
\midrule
\textsc{Baseline} & 18.2K & 56.3 \\
\textsc{Single} & 47.6K & 78.5 \\
\textsc{PER} & 401.9K & 84.9 \\
\textsc{Broadcast} & 622.1K & 88.9 \\
\bottomrule
\end{tabular}
\caption{Protocol-level cost and solve rate on the held-out test split.}
\label{tab:protocol-reference}
\end{table}

\begin{table}[ht]
\centering
\small
\begin{tabular}{lrr}
\toprule
Oracle label & Count & Fraction \\
\midrule
\textsc{Baseline} & 238 & 56.3 \\
\textsc{Single} & 97 & 22.9 \\
\textsc{PER} & 38 & 9.0 \\
\textsc{Broadcast} & 18 & 4.3 \\
\texttt{none} & 32 & 7.6 \\
\bottomrule
\end{tabular}
\caption{Fixed-order-oracle protocol distribution on the test split.}
\label{tab:oracle-distribution}
\end{table}

\subsection{Protocol Execution and Compute Accounting}
\label{app:execution-details}

All four protocol traces use the same underlying acting and evaluator model
family, \texttt{openai/gpt-oss-120b}, through an OpenAI-compatible inference
endpoint \citep{openai_gptoss_model}. The paper configs
use deterministic decoding with temperature 0.0 and a 4096-token per-call generation cap for solver/evaluator
calls. \textsc{Baseline} uses one direct solver attempt; \textsc{Single} adds
iterative self-correction with question-scoped repair memory; \textsc{PER}
uses planner, executor, and reviewer roles; and \textsc{Broadcast} uses
multi-agent deliberation with shared candidate state and peer approval. All
question-scoped memories and chat histories are reset between problems.

Token metrics in the routing tables are read from the matched trace logs. On
the full 4,181-problem benchmark, the four protocol runs account for about
4.53B logged tokens in total: 77M for \textsc{Baseline}, 201M for
\textsc{Single}, 1.67B for \textsc{PER}, and 2.58B for
\textsc{Broadcast}. The corresponding logged model-call totals are 40,501,
75,370, 416,037, and 562,939. These quantities are reported to make the
compute footprint of constructing the matched benchmark explicit; router-side
costs are included separately where applicable, such as in the confidence
cascade analysis. Logged average wall times per problem are about 44s, 74s,
4.7 minutes, and 4.9 minutes for the four protocols in the same order.

\subsection{Learned Router Implementation}
\label{app:learned-router}

The lightweight learned routers are five-class logistic regressions over the
same protocol labels used throughout the paper: Baseline, Single, PER,
Broadcast, and None. Text features are TF-IDF word unigrams and bigrams fit on
the training problems only
(\texttt{min\_df}=2, \texttt{max\_features}=20{,}000, lowercasing and Unicode
accent stripping). Metadata features include a one-hot source encoder, a
multi-label domain encoder, and standardized \texttt{difficulty} and
\texttt{difficulty\_tier} fields. The no-tier ablations remove both numeric
difficulty fields; the text-only router removes all metadata features.

For the main text+metadata and metadata-only routers, we train scikit-learn
\texttt{LogisticRegression} models with \texttt{solver=saga},
\texttt{random\_state=42}, and a grid over \(C \in \{0.25,1,4\}\) and
\texttt{class\_weight} in \{\texttt{None}, \texttt{balanced}\}. Validation
training uses warm-start one-epoch increments for at most 30 epochs and selects
the epoch with lowest dev log loss. The final configuration is selected on the
dev split by macro-F1, breaking ties by lower excess cost. The selected main
text+metadata model uses \(C=0.25\), no class weighting, and 15,124 total
features; the selected metadata-only model uses \(C=0.25\), no class weighting,
and 220 features.

\paragraph{Sentence-embedding router details.}
The embedding ablation uses only the problem statement text for the
embedding-only rows and concatenates those embeddings with the same source,
domain, difficulty, and difficulty-tier metadata features for the
embedding+metadata rows. We use a small frozen encoder rather than a fine-tuned
transformer so that the ablation remains a low-cost semantic-router check rather
than a new modeling contribution. Logistic variants sweep \(C \in
\{0.1,0.25,1,4,16\}\) with and without balanced class weights. kNN variants
sweep the neighbor count and distance weighting on the dev split. All reported
embedding-router results use the same train/dev/test split and the same
cost-aware evaluation code as the other router baselines.

\FloatBarrier

\section{Prompt and Execution Details}
\label{app:prompts}

\subsection{Frozen Router Prompt}

All frozen LLM routers receive the same metadata fields: problem source,
difficulty score, difficulty tier, domain summary, and full problem text. The
prompt gives the protocol order ordinally, not as exact token costs, so the
frozen router must infer the value of escalation from the problem and metadata
rather than from numeric token budgets. All frozen router runs use temperature
0.0. If a response cannot be parsed into one of the five allowed labels, the
offline evaluator assigns the fallback prediction \texttt{baseline\_llm}. To
reproduce the model call exactly, the listings retain the machine-readable
labels used during inference: \texttt{baseline\_llm}, \texttt{single\_agent},
\texttt{per}, \texttt{broadcast}, and \texttt{none} correspond to Baseline,
Single, PER, Broadcast, and None. The exact prompt is reproduced below.

\paragraph{System prompt.}
\begin{lstlisting}
You are a routing classifier, not a solver.
Do NOT solve the math problem.
Do NOT compute intermediate steps.
Your task is only to estimate the cheapest protocol likely to be sufficient.
Cost order from cheapest to most expensive: baseline_llm < single_agent < per < broadcast.
You may also output none if all protocols are unlikely to succeed.
Guidelines:
- baseline_llm: short direct calculations or standard manipulations.
- single_agent: moderate multi-step problems where one model with self-correction is likely enough.
- per: explicit planning and review are likely helpful.
- broadcast: especially hard problems that may benefit from multiple independent attempts.
Output exactly one label from: baseline_llm, single_agent, per, broadcast, none.
Start your reply with the label on the first line.
Stop after the label if possible.
\end{lstlisting}

\paragraph{User prompt template.}
\begin{lstlisting}
Problem source: {source}
Difficulty score: {difficulty}
Difficulty tier: {difficulty_tier}
Domain: {domain_summary}

Problem:
{problem}

Classify the likely cheapest sufficient protocol for this problem.
Do not solve the problem.
Reply with exactly one label on the first line: baseline_llm, single_agent, per, broadcast, or none.
\end{lstlisting}

\subsection{Cost-Aware Frozen Router Prompt Ablation}

The numeric-cost/few-shot ablation in Table~\ref{tab:embedding-cost-prompt}
uses \texttt{openai/gpt-oss-120b}, temperature 0.0, and the same test split as
main-paper Table~1. The parser accepts one of the five allowed labels
after an explicit \texttt{LABEL:} prefix; unparsable responses fall back to
Baseline. The final cleaned run has one fallback. The system
prompt below differs from the main frozen-router prompt only by adding numeric
average costs and short routing examples.

\paragraph{Cost-aware system prompt.}
\begin{lstlisting}
You are a routing classifier, not a solver.
Do NOT solve the math problem.
Do NOT compute intermediate steps.
Your task is only to estimate the cheapest protocol likely to be sufficient.
Cost order from cheapest to most expensive: baseline_llm < single_agent < per < broadcast.
You may also output none if all protocols are unlikely to succeed.
Guidelines:
- baseline_llm: short direct calculations or standard manipulations.
- single_agent: moderate multi-step problems where one model with self-correction is likely enough.
- per: explicit planning and review are likely helpful.
- broadcast: especially hard problems that may benefit from multiple independent attempts.
Output exactly one label from: baseline_llm, single_agent, per, broadcast, none.
Start your reply with the label on the first line.
Do not include reasoning unless the label already appears on the first line.

Approximate average token costs from the benchmark:
- baseline_llm: 18K tokens
- single_agent: 48K tokens
- per: 402K tokens
- broadcast: 622K tokens
Escalate only when the expected gain in solve probability justifies the extra token cost.

Routing examples:
- If a problem is routine arithmetic or a direct substitution, output baseline_llm.
- If a problem needs several algebraic steps but one solver is likely enough, output single_agent.
- If a problem likely needs explicit planning and checking, output per.
- If a problem is very hard and independent attempts may add value, output broadcast.
- If none of the four protocols is likely to solve it, output none.
\end{lstlisting}

\paragraph{Cost-aware user prompt template.}
\begin{lstlisting}
Problem source: {source}
Difficulty score: {difficulty}
Difficulty tier: {difficulty_tier}
Domain: {domain_summary}

Problem:
{problem}

Classify the cheapest protocol whose expected solve gain is worth its token cost.
Do not solve the problem.
Reply with exactly one line: LABEL: baseline_llm, LABEL: single_agent, LABEL: per, LABEL: broadcast, or LABEL: none.
\end{lstlisting}

\subsection{Direct Self-Assessment Prompt}

The headline self-confidence gate uses \texttt{gpt-oss-120b} because that
model produced the benchmark's Baseline solutions. The probe
asks for an explicit verbalized probability on a 0--100 scale in JSON; it does
not use token log-probabilities, self-consistency, or repeated sampling. The
main run uses the test split, temperature 0.0, a 256-token completion limit,
and at most 20 concurrent requests. Cross-model checks use the same JSON format but ask Llama-3.1-70B
or Gemma-3-27B to estimate \texttt{gpt-oss-120b}'s one-shot success; the
reduced Gemma actor check asks Gemma-3-27B about its own one-shot success on
that subset. Non-\texttt{gpt-oss} probes use at most seven concurrent requests.

\paragraph{Pre-answer confidence system prompt.}
\begin{lstlisting}
You are evaluating mathematical problem difficulty.
Do NOT solve the problem.
Do NOT compute intermediate steps.
If you start solving the problem, you are failing the task.
Your task is to estimate, without working through the solution, how likely you are to answer this problem correctly in a single direct attempt -- working alone, with no tools, no self-correction, and no collaboration.
Return only valid JSON with exactly this key:
{"SINGLE_PASS_PROB": <integer 0-100>}
Do not add any explanation, reasoning, or extra keys.
Your first character must be '{' and your last character must be '}'.
\end{lstlisting}

\paragraph{Pre-answer confidence user prompt template.}
\begin{lstlisting}
Problem source: {source}
Difficulty score: {difficulty}
Difficulty tier: {difficulty_tier}
Domain: {domain_summary}

Problem:
{problem}

Estimate your single-pass solve probability (0-100) without solving the problem.
Do not solve the problem.
\end{lstlisting}

For cross-model probes, the system and user prompts replace ``you'' with
``\texttt{openai/gpt-oss-120b}'' as the target solver. These runs are
diagnostic checks; they are not used to tune the main self-confidence gate.

\subsection{Protocol-Value Probe Prompt}

The exploratory protocol-value probe in Table~\ref{tab:protocol-value-probe}
uses the following JSON prompt.

\paragraph{Protocol-value system prompt.}
\begin{lstlisting}
You are estimating protocol success probabilities for math problem routing.
Do NOT solve the problem and do NOT compute intermediate steps.
Estimate how likely each protocol is to solve the problem correctly.
Protocol costs are approximately: baseline_llm 18K tokens, single_agent 48K, per 402K, broadcast 622K.
Return only valid JSON with exactly these integer keys, each from 0 to 100:
{
  "BASELINE_LLM_PROB": <integer>,
  "SINGLE_AGENT_PROB": <integer>,
  "PER_PROB": <integer>,
  "BROADCAST_PROB": <integer>
}
Do not add explanation, reasoning, or extra keys.
\end{lstlisting}

\paragraph{Protocol-value user prompt template.}
\begin{lstlisting}
Problem source: {source}
Difficulty score: {difficulty}
Difficulty tier: {difficulty_tier}
Domain: {domain_summary}

Problem:
{problem}

Estimate the probability that each protocol solves this problem correctly. Do not solve the problem.
\end{lstlisting}

\subsection{Pre-Answer Confidence Parsing and Coverage}

The initial parser recovered usable pre-answer confidence estimates for 310 of 423 test problems
(73.3\%). A manual re-parse of failed rows recovered 19 additional estimates,
yielding 329 usable rows (77.8\%). The remaining missing rows are not treated
as parser misses: 72 were truncated before an estimate was stated and 22 were
HTTP 429 failures. All pre-answer confidence figures and tables use the cleaned
analysis file and filter to rows with a recovered probability.

The same parse issue also appears on the dev split used for threshold tuning.
Most dev fallbacks were not network failures, but parse failures where the
model drifted into solving mode instead of returning the requested JSON
probability. We therefore do not treat missing confidence as missing at random.

We also tested a completion-cue variant that asked for a single integer. The
completion-cue variant was not used for the main analysis: a naive first-integer parser can silently
extract numbers copied from the problem statement rather than confidence
estimates. We therefore use the cleaned JSON-prompt run throughout the paper.
The coverage caveat matters for the interpretation of the mechanism results. The cleaning step
raises usable coverage but does not turn the probe into a routing oracle: even
after cleaning, low-confidence examples remain spread across multiple
fixed-order-oracle protocols and true failures.

\FloatBarrier

\section{Additional Diagnostic Figures}
\label{app:additional-figures}

\begin{figure*}[!t]
\centering
\begin{subfigure}{0.52\textwidth}
  \centering
  \includegraphics[width=\linewidth]{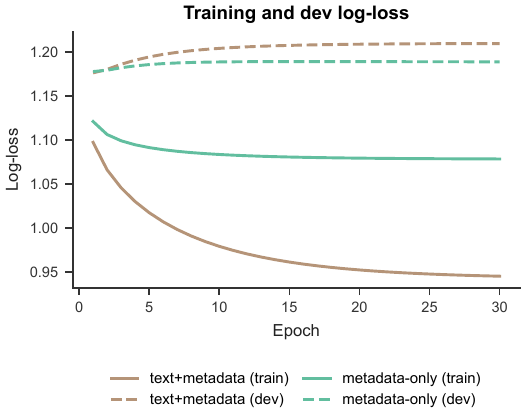}
  \caption{Training and dev log-loss. Text+metadata dev loss diverges from
  train loss after early epochs, motivating early stopping.}
  \label{fig:supp-loss}
\end{subfigure}

\vspace{0.8em}

\begin{subfigure}{0.92\textwidth}
  \centering
  \includegraphics[width=\linewidth]{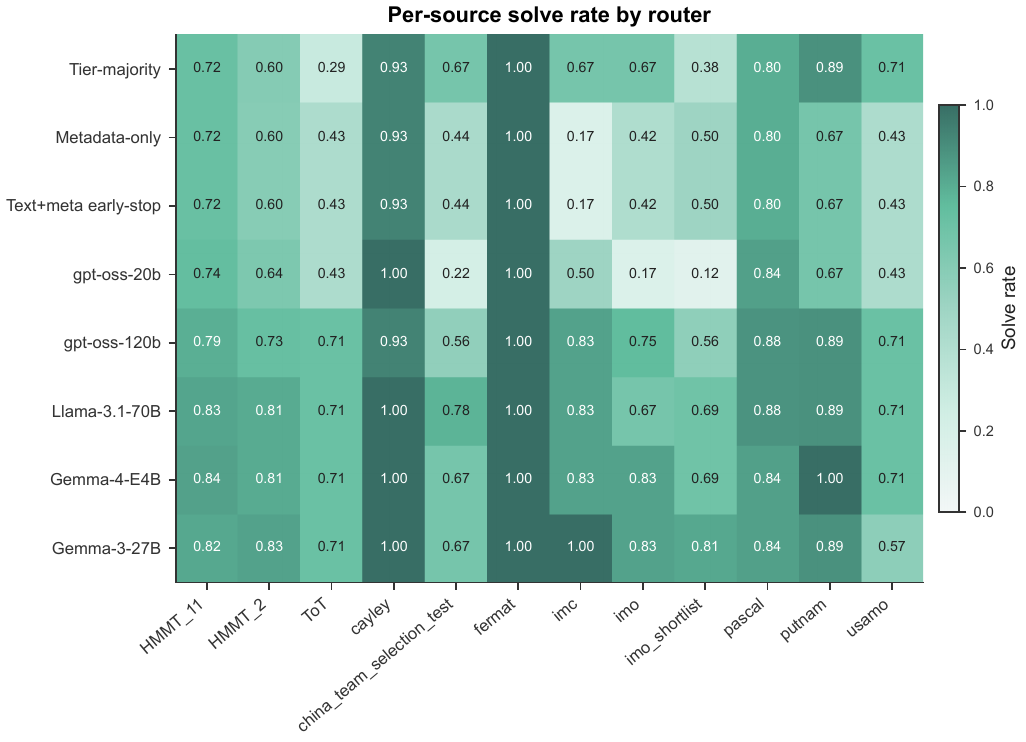}
  \caption{Per-source solve rates. The aggregate router ranking broadly
  persists across sources.}
  \label{fig:supp-source}
\end{subfigure}
\caption{Additional learned-router diagnostics. (a) The lightweight
text+metadata router begins to overfit after early epochs. (b) The aggregate
comparison is not explained by a single benchmark source.}
\label{fig:supp-diagnostics}
\end{figure*}

The final diagnostics support two boundary conditions for the main interpretation. The
loss-history plot checks that the learned-router result is not just a
single unlucky evaluation point: the text+metadata model continues improving on
training loss while its development loss worsens, consistent with overfitting
in this lightweight setup. The source-level heatmap checks that the router
comparison is not driven by one benchmark source alone. These diagnostics
support the narrower interpretation that the current routers are miscalibrated
about the marginal value of escalation.

\FloatBarrier

\section{Data and Reproducibility}
\label{app:reproducibility}

This appendix records the exact prompts, feature schemas, split rules,
execution settings, and uncertainty procedures needed to interpret the results.
The companion dataset archive is hosted at
\url{https://huggingface.co/datasets/ChihHsuan-Yang/scientific-agent-protocol-traces}.
For the settings studied here, it contains anonymized four-protocol traces,
per-protocol outcome labels, matched outcome tables, and registries that map
machine-readable identifiers to the Baseline, Single, PER, and Broadcast names
used in the paper. The arXiv source package also includes the aggregate result
tables underlying the reported comparisons as ancillary CSV files.

The companion archive covers additional experiments beyond this paper. The
claims in this paper use only the four protocols and the benchmark/model
settings defined in the main text; those boundaries should be used when
selecting records from the broader archive.

Raw problem redistribution follows each upstream benchmark's license; where
redistribution is restricted, the release provides identifiers and
reconstruction instructions rather than raw prompts.

\end{document}